\documentclass{article}

\usepackage{arxiv}

\usepackage[utf8]{inputenc} 
\usepackage[T1]{fontenc}    
\usepackage{hyperref}       
\usepackage{url}            
\usepackage{booktabs}       
\usepackage{amsfonts}       
\usepackage{nicefrac}       
\usepackage{microtype}      
\usepackage{amsmath}
\usepackage{lipsum}
\usepackage{graphicx}
\graphicspath{ {./images/} }
\usepackage{algorithm}
\usepackage{algpseudocode}
\usepackage{multicol}
\usepackage{multirow}
\usepackage{cellspace}
\usepackage{subcaption}
\usepackage[table]{xcolor}
\usepackage{booktabs}
\usepackage{multirow}
\usepackage{siunitx}
\usepackage{array}

\title{CARLE: A Hybrid Deep-Shallow Learning Framework for Robust and Explainable RUL Estimation of Rolling Element Bearings }

\author{
 Waleed Razzaq \\
  School of Automation\\
  University of Science and Technology China\\
  Hefei, Anhui \\
  \texttt{waleed.razzaq@mail.ustc.edu.cn} \\
   \And
 Yun-Bo Zhao \thanks{Corresponding author. Email: \texttt{ybzhao@ustc.edu.cn}} \\
  School of Automation\\
  University of Science and Technology China\\
  Hefei, Anhui \\
  \texttt{ybzhao@ustc.edu.cn} \\
}

\begin{document}
\maketitle

\setlength\cellspacebottomlimit{4pt} 
\setlength\cellspacetoplimit{4pt} 

\begin{abstract}
Prognostic health management (PHM) systems have extensive applications in industry for monitoring and predicting the health status of equipment. Remaining Useful Life (RUL) estimation stands out as one important part of a PHM system that predicts the remaining operational lifespan of mechanical systems or their components, such as rolling element bearings, which account for a high proportion of machinery failures. Although many methods for RUL estimation have been developed, there are some challenges in terms of generalizability and robustness under dynamic operating conditions. This paper introduces the CARLE AI framework, which integrates advanced deep learning architectures with shallow machine learning technique to overcome these limitations. CARLE integrates Res-CNN and Res-LSTM blocks with multi-head attention and residual connections to capture spatial and temporal degradation trends coupled with Random Forest Regression (RFR) for robust and accurate predictions. We further propose a compact feature extraction framework that implements Gaussian filtering for efficient noise reduction and Continuous Wavelet Transform (CWT) for time‒frequency feature extraction. We assessed the effectiveness of the proposed framework via the XJTU-SY and PRONOSTIA bearing datasets. Ablation experiments were conducted to assess the contribution of each component within CARLE, whereas noise experiments evaluated its resilience to noise. Cross-domain validation experiments were performed to examine the model's generalizability across multiple domains. Additionally, comparative analyses with several state-of-the-art methods under dynamic operating conditions demonstrated that CARLE outperformed competing approaches, particularly in terms of generalizability to unseen scenarios. Furthermore, we discuss the reliability and trustworthiness of this framework via multiple state-of-the-art explainable AI (XAI) techniques, i.e., LIME and SHAP.
\end{abstract}

\keywords{Prognostics Health Management (PHM) \and Remaining Useful Life (RUL) \and CNN \and LSTM \and XAI }

\section{Introduction}

Prognostic Health Management (PHM) systems play crucial roles in industries as they monitor and predict equipment health conditions to prevent severe operational safety hazards and ensure accident-free processes. One salient feature of PHM systems is Remaining Useful Life (RUL) estimation, which concentrates on estimating the remaining effective lifespan of machinery or its components. Rotational machinery is more prone to failure because of the availability of rolling-element bearings working under aggressive environments. It has been estimated that 40 to 50$\%$ of machinery failures can be attributed to these bearings \cite{zhuang2021temporal}. Therefore, an accurate RUL estimation system for rolling-element bearings is essential for monitoring degradation, mitigating risks, and preventing unexpected breakdowns. Recently, various methods have been developed for this purpose and can generally be divided into physics-based and data-driven models.

Physics-based models provide insights into the degradation processes of bearings via a set of equations derived from mathematical representations of physical systems. Guo et al. \cite{guo2015fatigue} proposed a physics-based model for bearing degradation based on Hertzian contact theory and material fatigue that effectively predicts nonlinear degradation under varying operational conditions. Wu et al. \cite{wu2016review} proposed a model with elastic deformation and stress distribution in ball bearings for simulating the initiation and development of spalls to show the merits of contact mechanics in understanding the early evolution of faults. Although these methods have achieved notable accomplishments, they require broad interdisciplinary knowledge and depend upon complicated mathematical modeling.

Data-driven methods uncover the hidden relationships within condition monitoring data. Further, it can be divided into two subcategories: shallow machine learning and deep learning. Bienefeld et al. \cite{bienefeld2022importance} explored Radom Forest (RF) performance in RUL estimation of rolling-element bearings using an extended feature engineering strategy involving the time domain, frequency domain, and statistical features extracted from vibrational signals. Zhang et al. \cite{zhang2021rotating} proposed a Relevance Vector Machine RVM-coupled method that integrates the advantages of health indication fusion to create one unified health indicator out of a set of vibrational and temperature-motivated features. The number of developments in monitoring data acquisition continues to increase significantly, making meaningful feature extraction of monitored multisensory data even more crucial for RUL estimation. However, most shallow machine learning algorithms have notable limitations in dealing with big data in terms of prediction accuracy and computational efficiency.

Deep learning architectures are designed to capture and represent rich patterns in big data through the composition of a neural network made of multiple hidden layers composed of perceptrons. Advanced deep learning algorithms, including CNN\cite{alzubaidi2021review}, recurrent networks such as LSTM \cite{hochreiter1997long} and GRU \cite{chung2014empirical}, and attention mechanisms \cite{vaswani2017attention} have proven highly efficient in uncovering hidden relationships within big data learning for RUL estimation of rolling element bearings. Li et al. \cite{li2019understanding} proposed a CNN-based approach using vibrational signal spectrograms and demonstrated very good performance, thus proving its ability to learn nonlinear degradation trends distinguishing subtle data variations in data. However, CNNs struggle to model temporal degradation trends and long-term time dependencies within big data, limiting their real-world applicability. Zhang et al. \cite{zhang2018long} utilized an LSTM-based network that effectively models long-term dependencies and captures temporal degradation features within massive datasets; however, its sensitivity to hyperparameters, overfitting and lack of noise handling limit its accuracy. Li et al. \cite{li2023remaining} proposed a GRU-based DeepAR network that was efficient in modeling temporal dependencies with parameters and an adaptive failure threshold. However, it is sensitive to noise and often requires careful tuning in complex cases. Deng et al. \cite{deng2023calibration} presented a calibrated hybrid transfer learning framework including a dynamic rolling bearing model, particle filter-based calibration, and a physics-informed Bayesian deep dynamic network for improving fidelity. However, it is still computationally intensive and has limited applicability in real-world conditions. Zhao et al. \cite{zhao2023multi} proposed Multiscale Integrated Self-Attention that performs with multisensory degrading data at various scales by employing a multiscale CNN block including a self–attention mechanism, a recurrent network module and feature fusion to extract multisensory-temporal features on the basis of their relationships and integrate them via mutual interaction. Although this approach improves prediction accuracy through an efficient loss function, it is hindered by varying sensor quality and data noise.

In addition to the individual limitations mentioned above, several other common challenges demand attention. Most of the methods reported in the literature are task-oriented, diminishing their real-world applicability for many industrial machinery operations where real conditions are highly variable. The second significant limitation concerns the generalizability and robustness of RUL prediction systems, which heavily depend on effective feature extraction. Most existing approaches do not have a robust and compact framework for feature engineering; hence, they have limited reliability when dealing with big data. Another limitation concerns the fact that they are not transparent. Predictions are given in a black-box way, without underlining any factors of rationale that may contribute to supporting such an outcome. Therefore, the inability of the data-driven RUL model to offer interpretability or explainability raises concerns regarding dependability and trust.

Given these drawbacks, we propose a causal RUL estimation system that learns from one working condition and generalizes its learning to others. We aim to achieve this goal by designing a compact feature extractor framework that accounts for noise and provides a concise feature vector for the AI system. For the AI system, we introduce CARLE (Deep Ensemble Residual Convolutional-Attention LSTM Network) consisting of four distinct network blocks: \textit{Res-CNN block}, \textit{Res-LSTM block}, \textit{Linear block} and \textit{ML block}. The Res-CNN block comprises several convolutional layers that extract spatial degradation trends from the input vector. These features are passed to a multi-head attention mechanism (MHA) that selects the most relevant spatial features, suppresses redundant features, and enables differential treatment of features by scanning global information. The output is subsequently fed into the Res-LSTM network to capture temporal dependencies and long-term relationships between features. Residual connections between the CNN and LSTM layers are introduced to increase the robustness and generalizability of the system while also easing the computational complexity associated with each architecture. Several linear layers are introduced in the Linear block to recognize patterns and generate a logit vector, which serves as input for the ML block that contains the Random Forest Regression (RFR) for the final prediction. We validate the performance of the system on the XJTU-SY and PRONOSTIA bearing datasets. We also discuss the trustworthiness of the AI framework via state-of-the-art explainable AI (XAI) techniques called LIME and SHAP, which allow us to assess whether the output prediction is reliable. The highlights of this research are listed below.
\begin{enumerate}
    \item A compact time-frequency feature extraction framework is designed to handle noise via a Gaussian filter and to extract diverse features from multichannel sensory data in both the time and frequency domains using Continuous Wavelet Transform (CWT).
    \item A novel CARLE AI system is designed for rolling-element bearings RUL estimation. The system ensemble the pattern-learning strength of multiple deep-learning architectures with the generalizability and robustness of shallow machine-learning algorithm.
    \item The effectiveness of the algorithm is validated on the XJTU-SY and PRONOSTIA bearing degradation datasets, which include data from multiple operating conditions.
    \item The reliability and trustworthiness of the proposed black-box framework are analyzed through multiple state-of-the-art XAI techniques, i.e., LIME and SHAP.
\end{enumerate}
The remainder of the paper is organized as follows: Section 2 outlines the methodology and algorithms utilized in the research. Section 4 presents the experimental results and analysis of the proposed framework. Section 5 concludes the research by discussing potential future work and areas for improvement. The appendix presents an overview of the foundational elements of the proposed framework, including a discussion on the feature extraction algorithms and training setup with hyperparameters of our implementation.

\begin{figure}[ht!]
\centering
\includegraphics[width=0.5\textwidth]{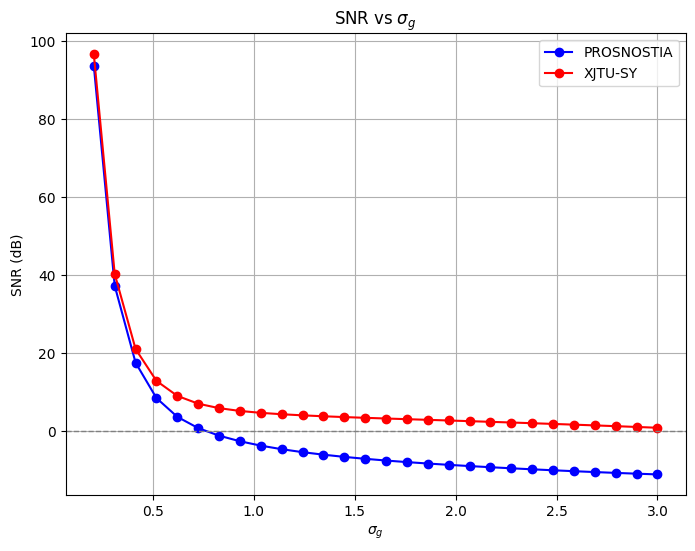}
\caption{Signal-to-Noise ratio (SNR) analysis w.r.t $\sigma_g$. The analysis shows the balance between noise reduction and signal preservation. When no smoothing is applied, the SNR remains high due to preservation of the original signal’s fidelity. As the smoothing parameter increases, the filtering mechanism effectively reduces high-frequency noise. However, the process simultaneously diminishes the finer details and dynamic components of the signal, resulting in a rapid decline in the SNR. After a certain smoothing intensity, noise reduction occurs at the cost of negligible signal distortion ($\sigma_g \approx 0.75$ for XJTU-SY and $\sigma_g \approx 1.1$ for PRONOSTIA). This stabilization point signifies an optimal parameter range where the balance between noise suppression and signal integrity is achieved.}
\label{fig:snr}
\end{figure}

\section{Time‒Frequency Feature Extraction Framework}

\begin{figure}[h!]
\centering
\includegraphics[width=0.85\textwidth]{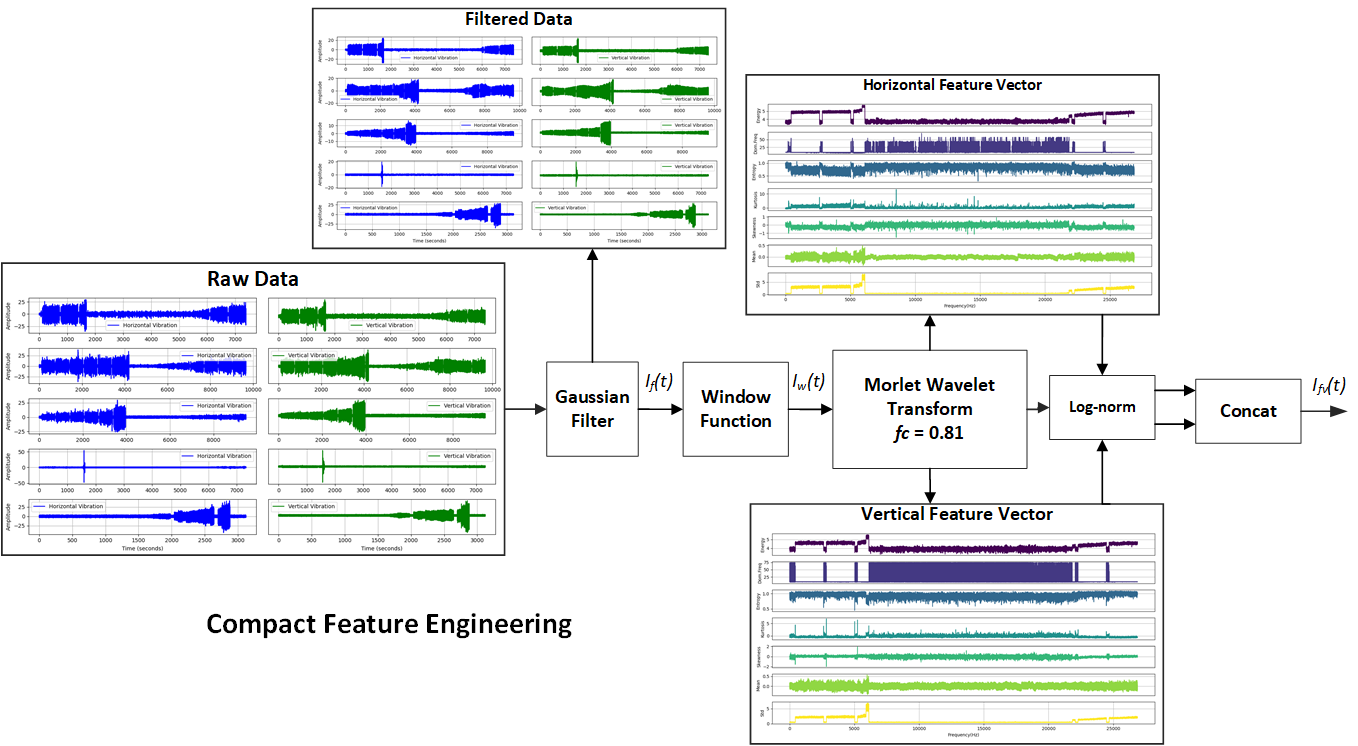}
\caption{Schematic diagram of the compact feature extractor framework.}
\label{fig3}
\end{figure}

The monitoring data from rolling element bearings typically consist of signals from multiple sensors, often exhibiting time-varying characteristics with perturbations, primarily thermal noise caused by changing operating conditions and temperature variations. To ensure effective analysis, a compact feature engineering preprocessing step is essential to filter out these perturbations before training the AI system (see algorithm \ref{algo1}). Otherwise, the perturbations can significantly degrade the performance of the system. Assuming that the number of sensors is \(N_s \) and the data length is \(L_d \), the raw data are represented as:
\begin{equation}
    I = [i_1, i_2, \dots, i_{L_d}], \quad i_k = [n_k^1, n_k^2, \dots, n_{N_c}^k]
\end{equation}

where \(i_k \) represents the sensor reading at timestep \(k \) for \(N_c \) channels. We filtered out the obtained raw data via Gaussian filter to smooth the edges and reduce short-term fluctuations. The choice of Gaussian filter is motivated by its effectiveness in reducing Gaussian noise while preserving signal edges, offering computational efficiency and reliable smoothing compared to the Fourier transform \cite{bracewell1989fourier} or EMD methods \cite{stallone2020new}. Let the filtered signal be denoted as \(I_f(t) \). The smooth signal is obtained by convolving the raw signal with the Gaussian filter ($G(x)$) (see Appendix A), represented mathematically as:
\begin{equation}
   I_f(t) = \int_{-\infty}^\infty I(\tau)G(t-\tau) \, d\tau
\end{equation}
The value of the standard deviation \(\sigma_g \) for the Gaussian filter plays a critical role in this process. It controls the degree of smoothing applied to the signal. After experimenting with various values and analyzing the signal-to-noise (SNR) (see Figure \ref{fig:snr}), we find the optimal balance for our use cases, effectively filtering out noise and thermal perturbations while preserving critical information essential for RUL estimation. The filtered signal \(I_f(t) \) is then forwarded to the CWT (see Appendix A) for feature extraction. However, before applying the CWT, the signal is divided into smaller segments using a windowing technique. The window operation can be represented as:
\begin{equation}
i_w(t) = I_f(t) \cdot w_i(t)
\label{eq:windowed}
\end{equation}
where \(w_i(t) \) is the window function with window length \(T_w \) for the \(i \)-th segment, defined as:
\begin{equation}
    w_i(t) =
\begin{cases}
1 & \text{if } t \in [t_i, t_{i+1}+T_w] \\
0 & \text{otherwise}.
\end{cases}
\label{eq:initi_window}
\end{equation}
By breaking down the signal into smaller segments, the CWT ensures that localized time‒frequency features are captured, which is vital for accurately modeling degradation trends for accurate RUL estimation. The CWT can then be mathematically computed as:
\begin{align}
\Gamma_{iw}(a, b) = \int_{-\infty}^\infty i_w(t) \psi^* \left(\frac{t - b}{a} \right) dt
\label{eq:cwt}
\end{align}
where \(\Gamma_{iw}(a, b) \) represents the wavelet coefficients of the windowed signal and where $\psi$ is the Morlet wavelet. To extract meaningful features from the CWT, it is critical to carefully select the frequency range of interest \((f_\text{min}, f_\text{max})\), as this range defines the scale range of the CWT. The choice of these frequencies is informed by the system’s operational condition \(f_o \), allowing the model to accommodate multiple scenarios effectively. In our implementation, we considered up to the third harmonic, providing a good balance between computational efficiency and capturing useful features. The frequency bounds are as follows:
\begin{equation}
   f_\text{min} \approx \frac{f_o}{3}, \quad f_\text{max} \approx 3f_o
   \label{eq:min_max}
\end{equation}
The corresponding wavelet transform scales can be calculated as:
\begin{equation}
  a_\text{min} = \frac{f_c}{f_\text{max}\cdot T_\text{sampling}} , \quad
a_\text{max} = \frac{f_c}{f_\text{min}\cdot T_\text{sampling}}
\label{eq:a_min_max}
\end{equation}
where \(T_\text{sampling} = 1/f_\text{sampling} \) is the period of the sampled vibrational signal and \(f_c \) is the central frequency of the Morlet wavelet, typically chosen as \(f_c = 0.81 \), to govern the trade-off between time and frequency resolutions. To ensure comprehensive coverage of the frequency range, logarithmically spaced scales are used:
\begin{equation}
a_i \in [a_\text{min}, a_\text{max}], \quad i = 1, 2, \dots, N
    \label{eq:log_a}
\end{equation}
where \(N \) is the number of scales selected on the basis of the desired resolution in the time‒frequency domain. Figure \ref{fig3} presents the visual representation of the compact feature extractor framework. The following time‒frequency representation (TFR) features are derived to characterize the system's physical state:

\begin{itemize}
    \item  \textbf{Energy (\( E \)):} represents the vibrational activity of the system. A continuous increase in energy typically correlates with progressive wear or distributed fatigue within the system, often evident as surface pitting. In contrast, sudden spikes indicate localized defects, such as spalling or crack propagation \cite{randall2011rolling, smith2015rolling}. Lubrication failures contribute to significant fluctuations, primarily due to the occurrence of intermittent metal-to-metal contact, whereas contamination, such as ingress of debris, results in transient energy spikes. The energy is computed as:
    \begin{align}
    E = \sum_{m=1}^{M} \left| \Gamma_{i_w}(a,b) \right|^2
    \label{eq:energy}
    \end{align}

    \item \textbf{Dominant frequency (\(f_d \)):} corresponds to the frequency at which the systems exhibit the highest energy concentration. Shifts in \(f_d \) can serve as a diagnostic tool for identifying specific faults within the system. Alignments with bearing fault frequencies, such as the ball pass frequency, are indicative of localized defects, commonly in the form of inner or outer race cracks (BPFO/BPFI) \cite{borghesani2013application, tandon1999review}. The presence of subharmonic components in \(f_d \) suggests potential issues such as looseness or imbalance within the system. Broadband frequency-domain profiles are characteristics of chaotic faults, which are typically associated with lubrication failures or contamination, as they introduce fluctuations in the system’s behavior. The dominant frequency is calculated as:
    \begin{align}
    f_d = a_{\text{scale}}(\text{\text{argmax}}(E))
    \label{eq:fd}
    \end{align}

    \item \textbf{Entropy (\(h \))}: measures the vibrational randomness within the system. Elevated entropy values suggest non-stationary defects, such as irregular spalling or looseness. In the case of lubrication failure, the entropy increases due to erratic friction, while corrosion-related damage leads to increased entropy through surface interactions. Early-stage fatigue typically indicates low entropy, which escalates as the degradation process becomes more chaotic. The entropy is calculated as:

    \begin{align}
    h = -\sum_{i=1}^{K} P(i_w) \log P(i_w)
    \label{eq:entropy}
    \end{align}

    \item \textbf{Kurtosis (\(K \)):} detects transient impacts by analyzing extreme deviations in the signal distribution. The highest kurtosis values are typically associated with localized defects, including fatigue cracks, electrical pitting, and particle collisions caused by contamination \cite{antoni2006spectral}. Kurtosis is calculated as follows:
    \begin{align}
    K = \frac{\mathbb{E}[(i_w - \mu)^4]}{\sigma^4}
    \label{eq:kurtosis}
    \end{align}

    \item \textbf{Skewness (\(s_k \)):} measures the asymmetry in the distribution of signal data. Positive skewness typically indicates unidirectional impacts, such as brinelling, while negative skewness suggests repetitive low-energy events, like the initiation of cracks. Asymmetric wear patterns resulting from thermal warping or corrosion also manifest as deviations in skewness, highlighting an imbalance in the system's behavior. The skewness is calculated as:
    \begin{align}
    s_k = \frac{\mathbb{E}[(i_w - \mu)^3]}{\sigma^3}
    \label{eq:skewness}
    \end{align}

    \item \textbf{Mean (\(\mu)\):} The mean vibrational level serves as a baseline indicator of system behavior. A gradual increase in the mean is often associated with distributed wear processes, such as corrosion or thermal degradation, whereas a sudden shift typically signals more severe faults, such as cage features. Lubrication failures can elevate the mean due to an increase in friction in the system. The mean is calculated as:
    \begin{align}
    \mu = \frac{1}{N} \sum_{m=1}^{M}i_w(m)
    \label{eq:mean}
    \end{align}

    \item \textbf{Standard deviation (\(\sigma \)):} represents the variability of the signal. High values indicate unstable faults such as looseness or contamination, which cause erratic behavior. Conversely, fatigue cracks contribute to increased variability during intermittent spalling events, indicating ongoing damage and instability in the system. The standard deviation is calculated as:
    \begin{align}
    \sigma = \sqrt{\frac{1}{M} \sum_{i=1}^{m} (i_w(m) - \mu)^2}
    \label{eq:std}
    \end{align}
\end{itemize}

While many existing approaches \cite{lei2007fault, lu2024physics,abdellatief2025ai, abdellatief2025comparative} use 20 or more TFR features, we extract only seven physically meaningful features, reducing offline computation time by on average 66\%. Integrating these features, such as transient detection through \(K \) and \(h \), with long-term trend analysis via \(\mu \) and \(\sigma \) can enhance RUL estimation. An increase in \(\mu \) with intermittent spikes in \(K \) indicates progressive wear punctuated by transient damage events. This allows for adaptive RUL updates that account for both ongoing wear and irregular fault occurrences. Similarly, the chaotic behavior observed in \(s_k \) and shifts in \(f_d \) improve prognostic accuracy by isolating fault-specific degradation pathways, allowing for a more precise RUL.

\begin{algorithm}[h!]
\caption{Time‒frequency Feature Extraction Framework}\label{algo1}
\begin{algorithmic}[1]
\Require Non-stationary vibrational signal ($I(t)$), $\sigma_g$, number of sensors ($N_s$), window length ($T_w$), sampling frequency ($f_s)$,
\State Initialize Gaussian filter:  $G(t) = \frac{1}{\sqrt{2\pi \sigma_g}} e^{-\frac{t^2}{2\sigma_{g}^2}}$
\State Calculate filtered signal:  $I_f(t) = \int_{-\infty}^\infty I(\tau)G(t-\tau) \, d\tau$
\State Initialize window function: $w(t) \gets$ Equation \ref{eq:initi_window}
\State Initialize central frequency: $f_c = 0.81$
\State $a_\text{min},a_\text{max} \gets$ Equation \ref{eq:a_min_max}
\State $a_{scale} \gets$ Equation \ref{eq:log_a}

\For{$n =1 \ldots N_s$}
    \For{$k =1 \ldots len(I_f(t))-T_w$}
        \State Compute window signal: $i_w(k) \gets$ Equation \ref{eq:windowed}.
        \State Compute wavelet coefficients: $\Gamma_{i_w}(a,b) \gets$ Equation \ref{eq:cwt}
        \State Compute energy: $E_n \gets$ Equation \ref{eq:energy}
        \State Compute dominant frequency: $f_{d_{n}} \gets$ Equation \ref{eq:fd}
        \State Compute entropy: $h_n  \gets$ Equation \ref{eq:entropy}
        \State Compute kurtosis: $k_n \gets$ Equation \ref{eq:kurtosis}
        \State Compute skewness: $sk_n \gets$ Equation \ref{eq:skewness}
        \State Compute mean: $\mu_n \gets$ Equation \ref{eq:mean}
        \State Compute standard deviation: $\sigma_n \gets$ Equation \ref{eq:std}
        \State $I_{{fv}_{n}} \gets [\log(E_n), f_{d_{n}}, h_n, k_n, sk_n, \mu_n, \sigma_n ]$
	\EndFor
\EndFor
\State $I_{fv} = Concat(I_{{fv}_{1}}, I_{{fv}_{2}} \ldots I_{{fv}_{N_s}})$
\State \textbf{return} $I_{fv}$
\end{algorithmic}
\end{algorithm}

\section{CARLE Framework}

We propose CARLE (Deep Ensemble Residual Convolutional-Attention LSTM Network) for the accurate RUL estimation in rolling element bearings. Unlike stacking-based ensembles that primarily combine base learners \cite{ture2024stacking}, CNN-Bi-LSTM approaches designed around predictive maintenance policies \cite{wang2024dynamic}, or data fusion methods with stage division \cite{li2024ensembled}, CARLE integrates residual CNNs, attention-driven LSTMs, and Random Forest Regression into a single unified framework. This design preserves spatial-temporal degradation features and enhances adaptability to unseen operating conditions, providing broader generalization across diverse requirements. A schematic diagram of CARLE is shown in Figure \ref{fig:carle_block}. The CARLE architecture comprises four interconnected blocks:

\begin{figure}[ht!]
\centering
\includegraphics[width=0.75\textwidth]{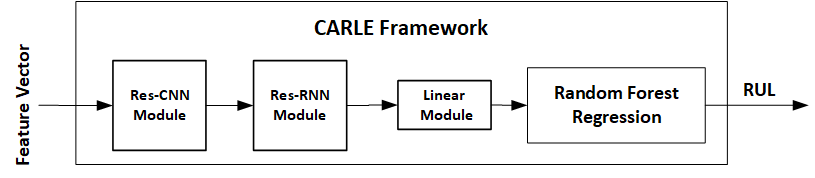}
\caption{The schematic diagram of the CARLE AI system.}
\label{fig:carle_block}
\end{figure}

\begin{figure}[ht!]
\centering
\includegraphics[width=0.9\textwidth]{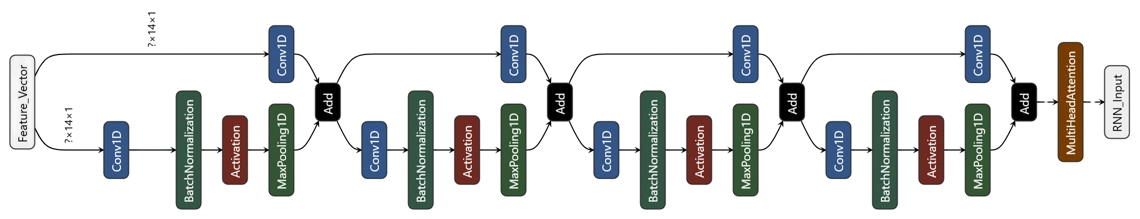}
\caption{Schematic diagram of the Res-CNN block.}
\label{fig:cnn_block}
\end{figure}

\textbf{Res-CNN Block}: receives the input feature vector ($I_{f_{v}})$ and processes it through multiple convolutional heads, each employing distinct filter and kernel sizes to extract salient degradation features. The MHA mechanism is incorporated at the output to enhance feature selection, allowing the model to prioritize relevant degradation features while minimizing redundant information. Additionally, residual connections are integrated to facilitate identity mapping, ensuring that vital features are retained and propagated throughout the network. This helps maintain accuracy in RUL predictions as the complexity increases. The schematic diagram of the Res-CNN is shown in Figure \ref{fig:cnn_block}.

\begin{figure}[ht!]
\centering
\begin{minipage}[b]{0.45\textwidth}
    \centering
    \includegraphics[width=\textwidth]{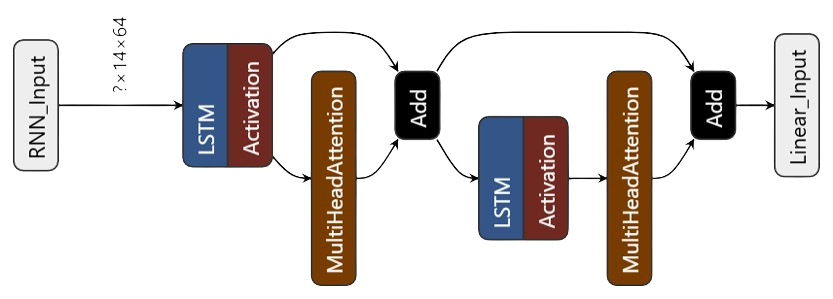}
    \caption{Schematic diagram of the Res-RNN block.}
    \label{fig:rnn_block}
\end{minipage}
\begin{minipage}[b]{0.45\textwidth}
    \centering
    \includegraphics[width=\textwidth]{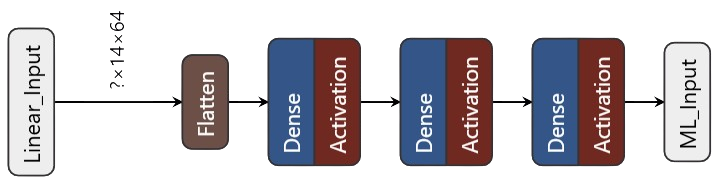}
    \caption{The schematic diagram of Linear block.}
    \label{fig:linear_block}
\end{minipage}
\label{fig:side_by_side}
\end{figure}

\textbf{Res-RNN Block}: receives the spatial degradation trends from the Res-CNN and processes them through a series of LSTM layers to capture the temporal characteristics and long-term dependencies inherent in the degradation features. Similar to the CNN block, a multi-head attention mechanism and residual connections are incorporated to enhance the focus on significant features and preserve critical information across layers. The schematic diagram of the Res-RNN is shown in Figure \ref{fig:rnn_block}.

\textbf{Linear Block}: consists of a series of fully connected layers tasked with recognizing patterns within the temporal degradation features, enabling the model to generalize effectively across diverse, unseen operating conditions. The output is a logit vector, which serves as input for the subsequent prediction mechanism. The schematic diagram of the Linear block is shown in Figure \ref{fig:linear_block}.

\textbf{Machine Learning Block}: The Random Forest Regression (RFR) model receives the logit vector from the linear block to enhance the generalization capabilities for new data, providing diverse perspectives and flexibility. RFR enhances generalization because it aggregates predictions from many decision trees trained on different subsets of the data and features. This ensemble averaging reduces overfitting, mitigates the effect of noise or outliers, and allows the model to capture diverse nonlinear relationships in the degradation features, making RUL predictions more robust to unseen operating conditions.

The stacking of these modules—CNN → Attention → LSTM → RFR—is deliberate. It reflects a layered processing approach: starting with low-level feature extraction, progressing to global pattern discovery, and concluding with structured temporal reasoning. This combination offers a comprehensive understanding of the degradation process, improving the robustness and accuracy of RUL predictions.

\section{Experimental Results and Analysis}

\subsection{Dataset Explanation}
\begin{figure}[ht!]
\centering
\includegraphics[width=0.9\textwidth]{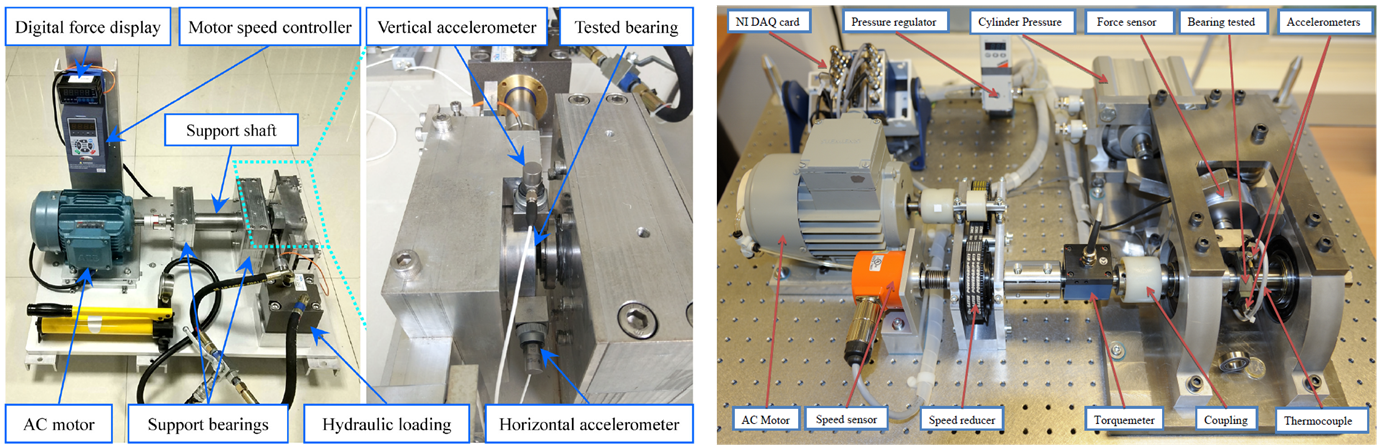}
\caption{\textbf{a)} XJTU-SY testbed; \textbf{b)} PRONOSTIA testbed for recording vibrational data}
\label{fig:testbed}
\end{figure}

\textbf{XJTU-SY dataset:} developed through a collaboration between Xi’an Jiaotong University and Changxing Sumyoung Technology for experimentation and validation of RUL algorithms \cite{wang2018xjtu}. The dataset includes run-to-failure vibration data from 15 rolling element bearings obtained through accelerated degradation experiments under three distinct operational conditions: 1200 rpm (35 Hz) with a 12 kN radial load, 2250 rpm (37.5 Hz) with an 11 kN radial load, and 2400 rpm (40 Hz) with a 10 kN radial load. Vibration signals were captured via accelerometers mounted on horizontal and vertical axes, sampled at a frequency ($f_{sample}$) of 25 kHz, and recorded at one-minute intervals, with each sample comprising 1.28 seconds of data. The experimental testbed is depicted in Figure \ref{fig:testbed}(a). For training, data from the $f_o=35 Hz$ condition (1200 rpm with a 12 kN load) were used, while validation focused on evaluating generalizability using data from the $f_o=40 Hz$ condition (2400 rpm with a 10 kN load) and $f_o=37.5$ condition (2250 rpm with an 11 kN load).

\textbf{PRONOSTIA dataset:} is a benchmark dataset widely used for research in condition monitoring and RUL analysis of rolling element bearings; it was developed as part of the PRONOSTIA experimental platform \cite{nectoux2012pronostia}. The dataset provides 16 complete run-to-failure data collected under accelerated degradation conditions with three distinct operational conditions: 1800 rpm (100 Hz) with a 4 kN radial load, 1650 rpm (100 Hz) with a 4.2 kN radial load, and 1500 rpm (100 Hz) with a 5 kN radial load. Vibration signals were captured via accelerometers mounted on the horizontal and vertical axes and sampled at 25.6 kHz, whereas temperature data were sampled at 10 Hz. Figure \ref{fig:testbed}(b) provides the testbed to capture the data. For training, we utilized 3 bearing data from 4KN operating conditions which is about 52$\%$ of total samples, and for validation, we focused on evaluating generalizability using data from 4.2 kN and 5 kN and ignored temperature data.

\subsection{RUL Labels}
\begin{figure}[ht!]
\centering
\includegraphics[width=0.5\textwidth]{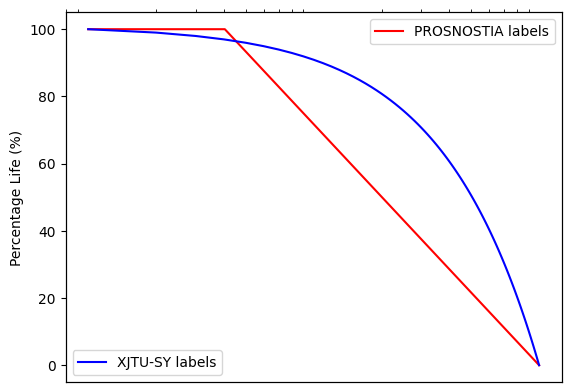}
\caption{RUL labels for both datasets.}
\label{fig:labels}
\end{figure}

Generating RUL labels is a crucial step in estimating remaining useful life. Some studies assume degradation occurs at a constant rate \cite{luo2022convolutional,yong2024remaining,deng2023calibration}, but real-world conditions rarely follow a perfectly linear pattern. Instead, degradation often occurs in a nonlinear, piecewise manner, as suggested in other studies \cite{zhao2023multi,yin2025physics,al2019multimodal}. To explore both possibilities, we created labels for the XJTU-SY dataset based on linear degradation models, visualized in Figure \ref{fig:labels} using a log scale for clarity. Since long-term monitoring data form a time series, the initial operation phase is typically stable, with minimal noticeable degradation. Therefore, for the PRONOSTIA dataset, we applied the nonlinear, piecewise degradation model shown in Figure \ref{fig:labels} to more accurately represent how bearing performance decreases over time.

\subsection{Evaluation indicators}
For evaluation, we utilized two metrics: the mean absolute error (MAE) and the root mean square error (MSE). A brief description of these metrics is as follows:

\textbf{MAE:} is widely used in RUL analysis to quantify the accuracy of predictive models. It measures the average magnitude of absolute errors between the predicted RUL ($y_i$) and the true RUL ($\hat{y}_i$), regardless of direction. The mathematical expression is as follows:
\[
\textit{MAE}= \frac{1}{n} \sum_{i=1}^{n} \left| y_i - \hat{y}_i \right|
\]
where $n$ is the total number of predictions. The MAE is particularly suitable for RUL analysis because it equally penalizes overpredictions and underpredictions, ensuring an unbiased evaluation of the model’s ability to estimate the RUL.

\textbf{MSE:} calculates the square root of the average squared differences between the predicted RUL ($y_i$) and the true RUL ($\hat{y}_i$). It is given by:
\[
\textit{MSE} = \sqrt{\frac{1}{n} \sum_{i=1}^{n} (y_i - \hat{y}_i)^2}
\]
Owing to the squaring of differences, the MSE penalizes larger errors more heavily. This makes it sensitive to significant prediction deviations, emphasizing the model’s ability to minimize large prediction errors.

\textbf{Score} is a metric specifically designed for RUL estimation in the IEEE PHM \cite{nectoux2012pronostia} to score the estimates. The scoring function is asymmetric and penalizes overestimations more heavily than early predictions. This reflects practical considerations, as late maintenance prediction can lead to unexpected failures with more severe consequences than early intervention can. 
\begin{equation}
\textit{Score} = \sum_{i : \hat{y}_i < y_i} \left( e^{-\frac{\hat{y}_i - y_i}{13}} - 1 \right) + \sum_{i : \hat{y}_i \geq y_i} \left( e^{\frac{\hat{y}_i - y_i}{10}} - 1 \right)
\end{equation}

\subsection{Ablation Experiments}

\begin{figure}[htbp]
\centering
\includegraphics[width=1.0\textwidth]{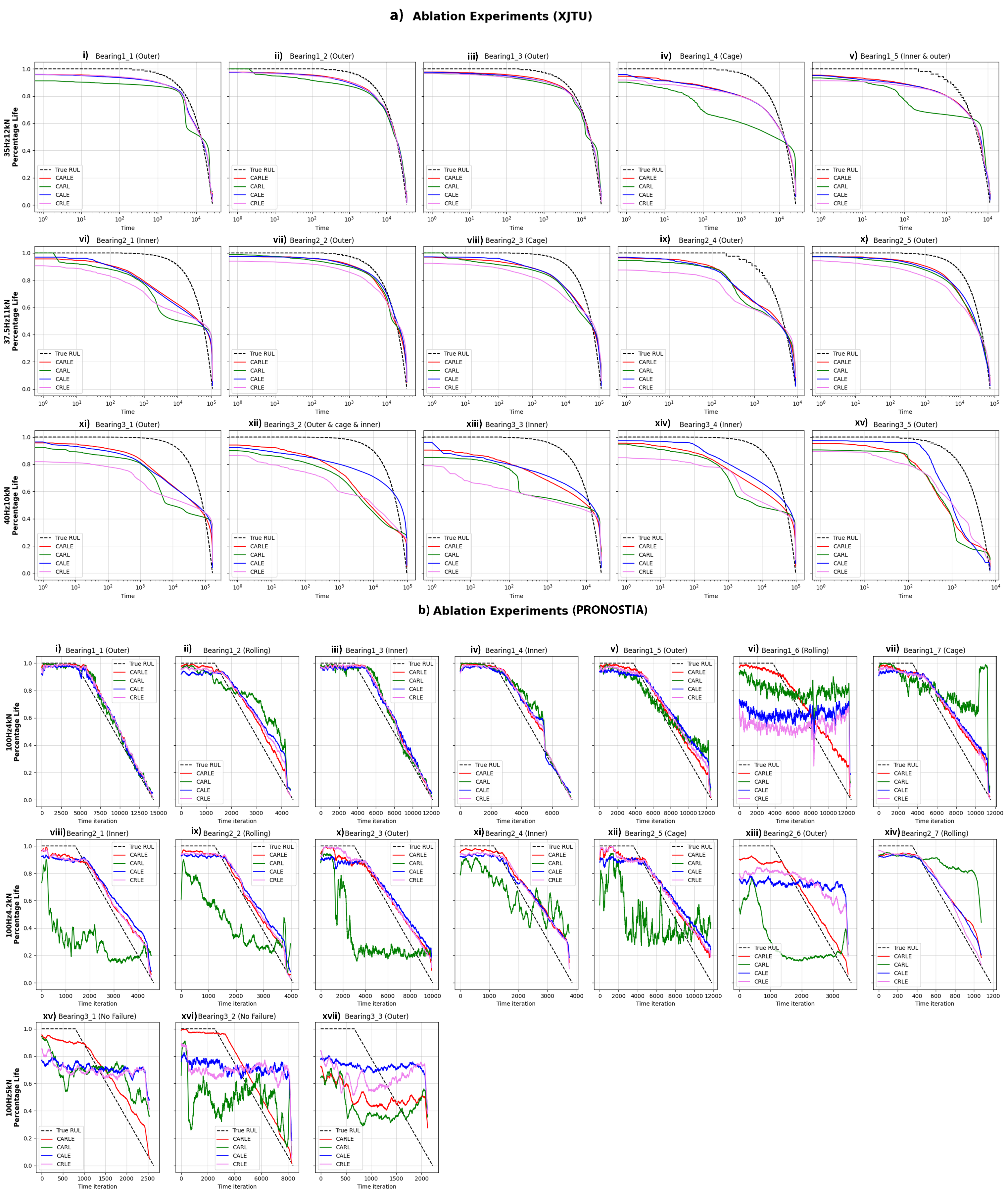}
\caption{Ablation experiment prediction \textbf{a)} XJTU-SY; \textbf{b)} PRONOSTIA with fault types.}
\label{fig:ablation_exp}
\end{figure}

Ablation experiments of CARLE were conducted to validate the effectiveness of each constituent of the architecture. We compared CARLE against its three variants: CARL without ensemble learning, CRLE without MHA, and CALE without residual connections. We noticed that CARLE and CALE performed very closely in terms of training operating conditions, but CARLE was marginally better than CALE. However, CARLE performed much better under unseen operating conditions, highlighting the role of residual connections in enhancing robustness. In contrast, both CRLE and CARL performed very poorly, with CARL being unsuitable for practical use. The ensemble machine learning approach yielded the most significant performance gains. A detailed comparison of both XJTU-SY and PRONOSTIA is shown in Figure~\ref{fig:ablation_exp}(a) and Figure~\ref{fig:ablation_exp}(b), while evaluation metrics are provided in Table~\ref{tab:ablation_xjtu} and Table~\ref{tab:ablation_pronostia}, respectively. For the sake of result explanations, we selected Bearing 3 under representative operating conditions from each operating condition and dataset.

For the XJTU dataset:

\begin{itemize}
    \item \textbf{35Hz12kN:} (Figure \ref{fig:ablation_exp}(a-iii)): CARLE achieved the lowest error with an MSE of 0.00220, MAE of 0.04087 and Score of 130.016. CALE followed closely with an MSE of 0.00265 (↑16\%), MAE of 0.04561 (↑10\%) and Score of 144.2532. CRLE recorded an MSE of 0.00275 (↑20\%), MAE of 0.04747 (↑13\%) and Score of 149.22, while CARL performed the worst with an MSE of 0.00806 (↑72\%), MAE of 0.07905 (↑48\%) and Score of 250.3705.
    
    \item \textbf{37.5Hz11kN:} (Figure \ref{fig:ablation_exp}(a-viii)): CARLE achieved an MSE of 0.01407, MAE of 0.10697 and Score of 1083.53. CALE showed slightly better MSE (0.01388, ↓1.3\%) but nearly identical MAE (0.10701, ↑0.03\%) with Score of 1081.08. CARL showed an MSE of 0.021 (↑33\%), MAE of 0.13195 (↑19\%) and Score of 1334.1443, while CRLE yielded an MSE of 0.02340 (↑39.87\%), MAE of 0.13731 (↑22\%) and Score of 1411.7924.
    
    \item \textbf{40Hz10kN:} (Figure \ref{fig:ablation_exp}(a-xiii)): CARLE maintained strong performance with an MSE of 0.03085, MAE of 0.15631 and Score of 331.6710. CALE demonstrated marginal improvements with an MSE of 0.02781 (↓9.8\%), MAE of 0.14869 (↓4.8\%) and Score of 323.87. Conversely, CARL and CRLE again exhibited degraded performance, recording MSEs of 0.05309 (↑42\%) and 0.05481 (↑43\%), MAEs of 0.20083 (↑22\%) and 0.20161 (↑22.4\%) and Score of 424.47 and 420.05, respectively.
\end{itemize}

For the PRONOSTIA dataset:

\begin{itemize}
    \item \textbf{100Hz4kN:} (Figure \ref{fig:ablation_exp}(b-iii)): CARLE achieved superior performance with an MSE of 0.00029 , MAE of 0.01312 and Score of60.912. CALE showed reduced accuracy with an MSE of 0.00094 (↑60\%), MAE of 0.02538 (↑48.3\%) and Score of 64.2970. CRLE performed moderately, with an MSE of 0.00049 (↑40\%), MAE of 0.01723 (↑23.8\%) and Score of 59.89, while CARL reached an MSE of 0.00033 (↑12.1\%), MAE of 0.01294 (↓1.2\%) and Score of 64.2970.
    
    \item \textbf{100Hz4.2kN:} (Figure \ref{fig:ablation_exp}(b-x)): CARLE achieved an MSE of 0.00831, MAE of 0.07488 and Score of 72.5470. CALE yielded an MSE of 0.01240 (↑32.9\%), MAE of 0.09776 (↑23.4\%) and Score of 96.5710. Interestingly, CRLE outperformed CARLE here, recording an MSE of 0.00601 (↓10\%), MAE of 0.04195 (↓56\%) and Score of 66.5046. CARL also showed strong results, with an MSE of 0.00601 (↓27.6\%), MAE of 0.03408 (↓54.4\%) and Score of 229.629.
    
    \item \textbf{100Hz5kN:} (Figure \ref{fig:ablation_exp}(b-xvii)): CARLE recorded an MSE of 0.14125, MAE of 0.17514 and Score of 37.2298. CALE improved significantly, with an MSE of 0.02628 (↓81\%), MAE of 0.14068 (↓19.6\%) and Score of 30.7763. CRLE achieved an MSE of 0.04916 (↓60\%), MAE of 0.17957 (↓2.4\%) and 25.7763, while CARL showed an MSE of 0.06594 (↓53\%) but a higher MAE of 0.22075 (↑26\%) with Score of 40.9659.
\end{itemize}

These findings confirm that each architectural component within CARLE makes a meaningful contribution to the overall model performance. Ensemble learning, in particular, drives substantial accuracy gains, while residual connections and attention mechanisms further support model generalization, especially in complex or unseen operational settings.

\begin{table}[htbp]
\centering
\caption{Ablation experiment (XJTU-SY)}
\vspace{3mm}
\label{tab:ablation_xjtu}
\footnotesize
\setlength{\tabcolsep}{2pt}
\renewcommand{\arraystretch}{1}
\resizebox{\textwidth}{!}{%
\begin{tabular}{@{}llSSSSSSSSSSS@{}}
\toprule
\rowcolor{gray!20}
\multirow{2}{*}{\textbf{Bearing}} & \multirow{2}{*}{\textbf{Model}} &
\multicolumn{3}{c}{\textbf{35Hz12kN}} &
\multicolumn{3}{c}{\textbf{37.5Hz11kN}} &
\multicolumn{3}{c}{\textbf{40Hz10kN}} \\
\cmidrule(lr){3-5} \cmidrule(lr){6-8} \cmidrule(lr){9-11}
 & & \textbf{MSE} & \textbf{MAE} & \textbf{Score} & \textbf{MSE} & \textbf{MAE} & \textbf{Score} & \textbf{MSE} & \textbf{MAE} & \textbf{Score} \\
\midrule

\multirow{4}{*}{\textbf{Bearing 1}}
 & CARLE & 0.00345 & 0.05157 & 122.5423 & \textbf{0.03273} & \textbf{0.16070} & \textbf{1505.8041} & \textbf{0.03314} & \textbf{0.15983} & \textbf{2269.5957} \\
 & CARL & 0.01377 & 0.09911 & 234.7118 & 0.05943 & 0.21274 & 1986.9069 & 0.06752 & 0.22609 & 3222.6052 \\
 & CALE & \textbf{0.00331} & \textbf{0.05126} & \textbf{121.5300} & 0.03630 & 0.16899 & 1581.9119 & 0.03954 & 0.17636 & 2564.5586 \\
 & CRLE & 0.00398 & 0.05639 & 133.9703 & 0.04990 & 0.19688 & 1865.2695 & 0.05716 & 0.20893 & 3030.7470 \\
\midrule

\multirow{4}{*}{\textbf{Bearing 2}}
 & CARLE & 0.00183 & \textbf{0.03692} & 114.4403 & \textbf{0.00671} & 0.07308 & \textbf{232.8929} & 0.06673 & 0.22032 & 1708.8516 \\
 & CARL & 0.00376 & 0.05666 & 175.5158 & 0.00883 & 0.08308 & 267.3460 & 0.07617 & 0.23716 & 1859.4258 \\
 & CALE & 0.00208 & 0.04082 & 125.9960 & 0.00786 & \textbf{0.06997} & 236.2859 & \textbf{0.02941} & \textbf{0.15117} & \textbf{1345.2411} \\
 & CRLE & \textbf{0.00178} & 0.03694 & \textbf{113.9618} & 0.01698 & 0.11337 & 368.1130 & 0.06164 & 0.21395 & 1672.4370 \\
\midrule

\multirow{4}{*}{\textbf{Bearing 3}}
 & CARLE & \textbf{0.00220} & \textbf{0.04087} & \textbf{130.0166} & \textbf{0.01407} & \textbf{0.10697} & \textbf{1083.5319} & \textbf{0.03085} & \textbf{0.15631} & \textbf{331.6710} \\
 & CARL & 0.00806 & 0.07905 & 250.3705 & 0.02100 & 0.13195 & 1334.1443 & 0.05309 & 0.20083 & 424.4746 \\
 & CALE & 0.00265 & 0.04561 & 144.2532 & 0.01388 & 0.10701 & 1081.0886 & 0.02781 & 0.14869 & 323.8718 \\
 & CRLE & 0.00275 & 0.04747 & 149.2223 & 0.02340 & 0.13731 & 1411.7924 & 0.05481 & 0.20161 & 420.0580 \\
\midrule

\multirow{4}{*}{\textbf{Bearing 4}}
 & CARLE & \textbf{0.01172} & \textbf{0.09653} & \textbf{225.5295} & \textbf{0.02149} & \textbf{0.12591} & 96.9172 & \textbf{0.03361} & \textbf{0.16221} & \textbf{1396.6897} \\
 & CARL & 0.05099 & 0.19865 & 461.9145 & 0.02965 & 0.14793 & 114.1304 & 0.06197 & 0.21669 & 1859.8220 \\
 & CALE & 0.01264 & 0.10096 & 236.8543 & 0.02065 & 0.11565 & \textbf{85.9529} & 0.03220 & 0.15930 & 1429.6296 \\
 & CRLE & 0.01332 & 0.10426 & 245.6415 & 0.03441 & 0.16022 & 124.7697 & 0.05450 & 0.20399 & 1745.6810 \\
\midrule

\multirow{4}{*}{\textbf{Bearing 5}}
 & CARLE & \textbf{0.00465} & \textbf{0.05938} & \textbf{59.9726} & 0.01373 & 0.09903 & 582.5476 & 0.09625 & 0.26167 & 154.0171 \\
 & CARL & 0.02127 & 0.12677 & 128.7710 & 0.01256 & 0.09288 & 547.5211 & 0.11684 & 0.28730 & 169.0601 \\
 & CALE & 0.00486 & 0.06090 & 61.3297 & \textbf{0.00985} & \textbf{0.08766} & \textbf{511.8520} & 0.08958 & 0.26006 & \textbf{151.7796} \\
 & CRLE & 0.00537 & 0.06420 & 64.8331 & 0.02060 & 0.12428 & 750.2275 & \textbf{0.06565} & \textbf{0.22405} & 130.8896 \\
\bottomrule
\end{tabular}}
\vspace{3mm}
\begin{minipage}{\textwidth}
\footnotesize
\vspace{3mm}
\textbf{Note:} Bold values indicate the minimum MSE, MAE, and Score for each bearing-condition combination.
\end{minipage}
\end{table}

\begin{table}[htbp]
\centering
\caption{Ablation experiment (PRONOSTIA)}
\vspace{3mm}
\label{tab:ablation_pronostia}
\footnotesize
\setlength{\tabcolsep}{2pt}
\renewcommand{\arraystretch}{1}
\resizebox{\textwidth}{!}{%
\renewcommand{\arraystretch}{0.5}
\begin{tabular}{@{}llSSSSSSSSSS@{}}
\toprule
\rowcolor{gray!20}
\multirow{2}{*}{\textbf{Bearing}} & \multirow{2}{*}{\textbf{Model}} &
\multicolumn{3}{c}{\textbf{100Hz4kN}} &
\multicolumn{3}{c}{\textbf{100Hz4.2kN}} &
\multicolumn{3}{c}{\textbf{100Hz5kN}} \\
\cmidrule(lr){3-5} \cmidrule(lr){6-8} \cmidrule(lr){9-11}
 & & \textbf{MSE} & \textbf{MAE} & \textbf{Score} & \textbf{MSE} & \textbf{MAE} & \textbf{Score} & \textbf{MSE} & \textbf{MAE} & \textbf{Score} \\
\midrule

\multirow{4}{*}{\textbf{Bearing 1}}
 & CARLE & \textbf{0.00017} & \textbf{0.00890} & 67.7651 & 0.00687 & 0.06874 & \textbf{34.2385} & \textbf{0.03073} & \textbf{0.15232} & \textbf{22.3816} \\
 & CARL & 0.00060 & 0.01944 & \textbf{52.1498} & 0.20664 & 0.39369 & 125.8891 & 0.01941 & 0.12100 & 26.8461 \\
 & CALE & 0.00130 & 0.02515 & 53.3133 & 0.01075 & 0.08844 & 44.6632 & 0.02536 & 0.13877 & 34.7183 \\
 & CRLE & 0.00055 & 0.01640 & 64.1635 & \textbf{0.00720} & \textbf{0.07223} & 34.3931 & 0.03786 & 0.17235 & 41.4252 \\
\midrule

\multirow{4}{*}{\textbf{Bearing 2}}
 & CARLE & 0.00289 & 0.04268 & \textbf{28.5949} & \textbf{0.00406} & 0.05340 & \textbf{22.4782} & \textbf{0.04126} & \textbf{0.17514} & \textbf{52.6417} \\
 & CARL & 0.01822 & 0.10301 & 53.7729 & 0.09584 & 0.26255 & 69.0456 & 0.06594 & 0.22073 & 114.1094 \\
 & CALE & 0.01010 & 0.08341 & 44.0762 & 0.00706 & 0.06948 & 33.5543 & 0.02629 & 0.14069 & 112.8656 \\
 & CRLE & \textbf{0.00643} & \textbf{0.06490} & 37.0388 & 0.00403 & \textbf{0.05156} & 26.3432 & 0.04196 & 0.17951 & 137.5010 \\
\midrule

\multirow{4}{*}{\textbf{Bearing 3}}
 & CARLE & \textbf{0.00029} & 0.01312 & 60.9126 & 0.00831 & 0.07488 & 72.5470 & 0.14125 & 0.17514 & 37.2298 \\
 & CARL & 0.00033 & \textbf{0.01294} & \textbf{53.8716} & 0.17432 & 0.34082 & 229.6209 & 0.065935 & 0.220728 & 40.9659 \\
 & CALE & 0.00094 & 0.02538 & 64.2970 & 0.01240 & 0.09776 & 96.5710 & \textbf{0.02628} & \textbf{0.14068} & 30.4960 \\
 & CRLE & 0.00049 & 0.01723 & 59.8920 & \textbf{0.00601} & \textbf{0.041959} & \textbf{66.5046} & 0.04916 & 0.17951 & \textbf{25.7763} \\
\midrule

\multirow{4}{*}{\textbf{Bearing 4}}
 & CARLE & \textbf{0.00052} & \textbf{0.01635} & \textbf{35.9291} & \textbf{0.01035} & \textbf{0.07616} & 39.0356 & \text{--} & \text{--} & \text{--} \\
 & CARL & 0.00237 & 0.03209 & 45.9774 & 0.02068 & 0.12315 & \textbf{34.4502} & \text{--} & \text{--} & \text{--} \\
 & CALE & 0.00268 & 0.04036 & 36.9779 & 0.01279 & 0.09542 & 38.6783 & \text{--} & \text{--} & \text{--} \\
 & CRLE & 0.00134 & 0.02676 & 37.9885 & 0.01029 & 0.08145 & 38.1909 & \text{--} & \text{--} & \text{--} \\
\midrule

\multirow{4}{*}{\textbf{Bearing 5}}
 & CARLE & \textbf{0.00264} & \textbf{0.03956} & \textbf{69.1269} & \textbf{0.00711} & \textbf{0.06936} & \textbf{91.5711} & \text{--} & \text{--} & \text{--} \\
 & CARL & 0.00885 & 0.07636 & 86.3878 & 0.08058 & 0.23612 & 178.1749 & \text{--} & \text{--} & \text{--} \\
 & CALE & 0.00999 & 0.08285 & 116.8831 & 0.01398 & 0.10261 & 122.0432 & \text{--} & \text{--} & \text{--} \\
 & CRLE & 0.00605 & 0.06274 & 94.7340 & 0.00878 & 0.07688 & 94.7391 & \text{--} & \text{--} & \text{--} \\
\midrule

\multirow{4}{*}{\textbf{Bearing 6}}
 & CARLE & \textbf{0.02014} & \textbf{0.12283} & \textbf{97.9291} & \textbf{0.07713} & \textbf{0.23869} & \textbf{24.9688} & \text{--} & \text{--} & \text{--} \\
 & CARL & 0.03497 & 0.13915 & 205.7567 & 0.20729 & 0.39659 & 98.0300 & \text{--} & \text{--} & \text{--} \\
 & CALE & 0.03064 & 0.15598 & 151.3578 & 0.03074 & 0.15201 & 51.7674 & \text{--} & \text{--} & \text{--} \\
 & CRLE & 0.05942 & 0.20851 & 184.1260 & 0.03314 & 0.15922 & 57.2092 & \text{--} & \text{--} & \text{--} \\
\midrule

\multirow{4}{*}{\textbf{Bearing 7}}
 & CARLE & \textbf{0.00799} & \textbf{0.06877} & \textbf{101.4600} & 0.00495 & \textbf{0.06176} & 12.3018 & \text{--} & \text{--} & \text{--} \\
 & CARL & 0.03480 & 0.13173 & 196.0093 & 0.06695 & 0.18207 & 26.0443 & \text{--} & \text{--} & \text{--} \\
 & CALE & 0.01221 & 0.08969 & 122.1526 & 0.00929 & 0.08194 & 11.2305 & \text{--} & \text{--} & \text{--} \\
 & CRLE & 0.00905 & 0.07806 & 108.2637 & \textbf{0.00355} & 0.05256 & \textbf{8.9416} & \text{--} & \text{--} & \text{--} \\
\bottomrule
\end{tabular}}
\vspace{3mm}
\begin{minipage}{\textwidth}
\footnotesize
\vspace{3mm}
\textbf{Note:} Bold values indicate the minimum MSE, MAE, and Score for each bearing-condition combination.
\end{minipage}
\end{table}

\subsection{Noise Experiment}

\begin{figure}[ht!]
\centering
\includegraphics[width=1.0\textwidth]{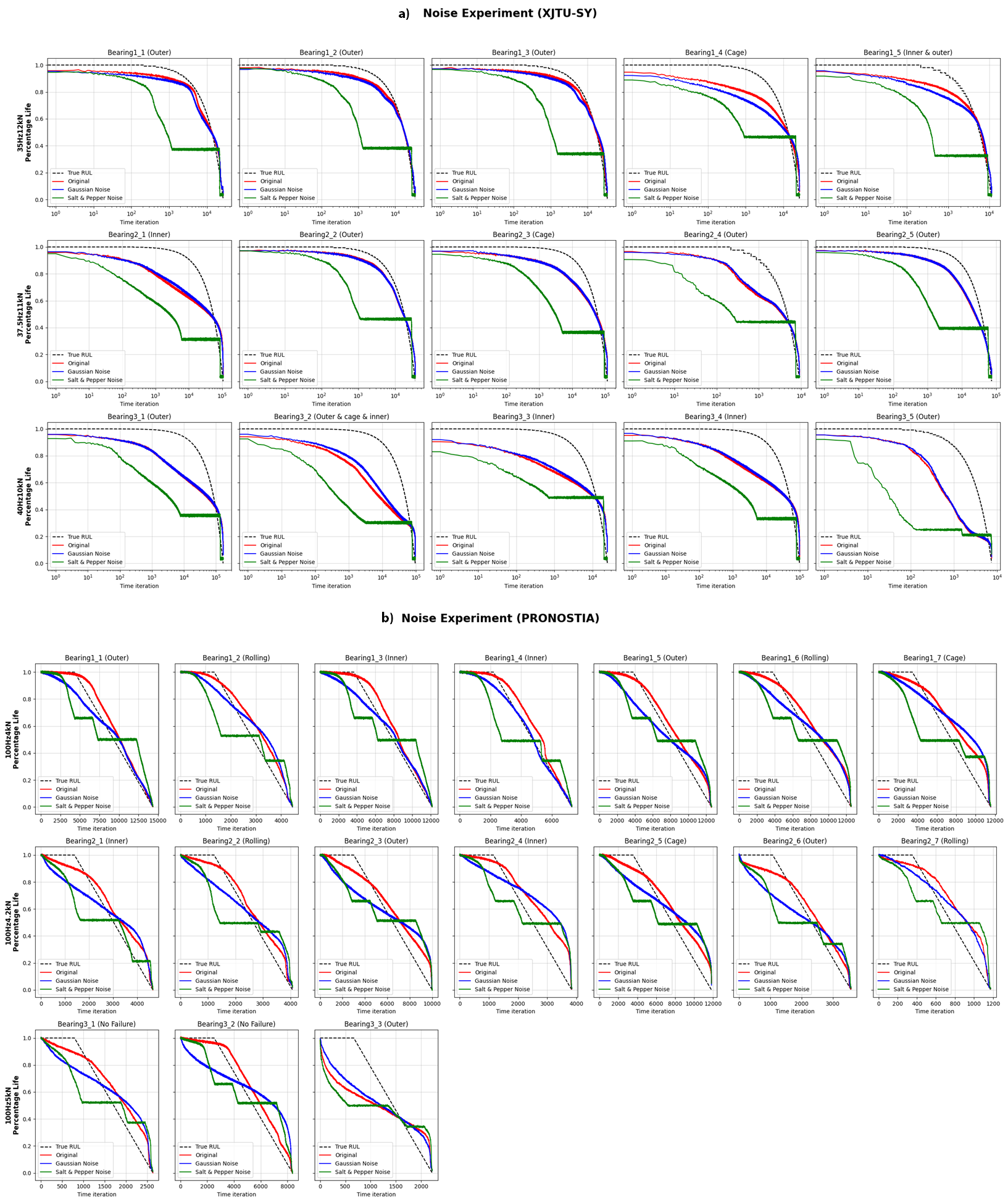}
\caption{Noise experiment result for \textbf{a)} XJTU-SY; \textbf{b)} PRONOSTIA.}
\label{fig:noise}
\end{figure}

Noise experiments are crucial for evaluating the robustness and reliability of AI frameworks, particularly in real-world scenarios where data are affected by sensor noise, environmental variations, or system uncertainties. By introducing controlled noise into the input data, we can assess the model’s stability and its ability to generalize beyond ideal conditions. In our experiments, Gaussian noise with a normal distribution ($\mu = 0$, $\sigma = 0.1$) was added to simulate typical sensor fluctuations. Additionally, salt-and-pepper noise was applied randomly to 10\% of the data points, representing sudden sensor failures. Results show that the model is largely resilient to Gaussian noise, with only minor performance degradation on the XJTU-SY dataset (Figure \ref{fig:noise}(a)). Salt-and-pepper noise, however, causes a more significant performance drop, highlighting a potential limitation for real-world deployment where sensor spikes or dropouts can occur due to electrical interference, hardware faults, or communication errors. In the PRONOSITA evaluation (Figure \ref{fig:noise}(b)), the impact of both noise types is more moderate, indicating that the model can still preserve long-term bearing degradation patterns. To mitigate the effect of salt-and-pepper noise in practice, preprocessing filters such as median or robust statistical filters can remove sudden spikes, sensor fusion can reduce the influence of any single faulty measurement, and training with noise-augmented data can help the model learn to ignore extreme outliers. Additionally, integrating lightweight anomaly detection modules could flag or correct extreme values in real time, ensuring more reliable RUL predictions under noisy conditions.

\subsection{Cross-domain Validation Experiments}
\begin{figure}[ht!]
\centering
\includegraphics[width=0.7\textwidth]{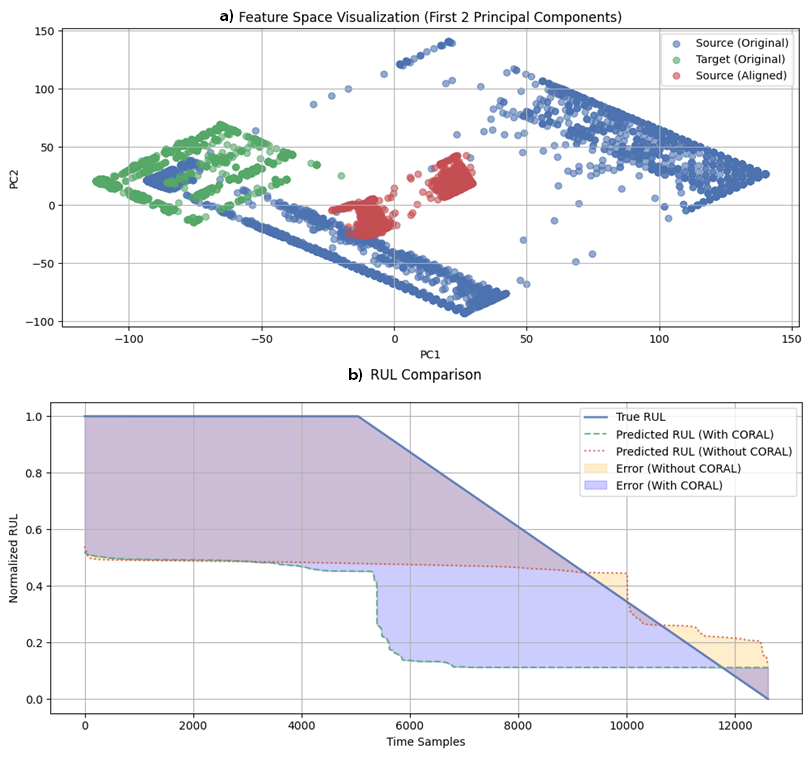}
\caption{\textbf{a)} Feature space alignment using CORAL-PCA; \textbf{b)} RUL comparison of CORAL-PCA-aligned and non-aligned.}
\label{fig:cross_domain}
\end{figure}

Cross-domain validation is crucial for assessing the generalizability of AI frameworks when applied to datasets with differing statistical distributions. It evaluates whether a model trained on one dataset can maintain predictive performance on another, thereby mitigating overfitting to a single domain and improving applicability in dynamic environments. We evaluate the PRONOSTIA-trained CARLE model on the XJTU-SY dataset, as both datasets share identical feature sets derived via Algorithm \ref{algo1} but differ in label distributions. To address domain shift, we employ Principal Component Analysis (PCA) and Correlation Alignment (CORAL) for feature space alignment. The process involves feature extraction from both datasets, transformation via PCA, and distribution alignment using CORAL (see Figure \ref{fig:cross_domain}(a)) before generating predictions. Our analysis (see Figure \ref{fig:cross_domain}(b) and Table \ref{tab:coral}) indicates that the adapted methodology produces varying prediction accuracy, with notable differences between CORAL-aligned and non-aligned results. Specifically, the CORAL-aligned model achieved an MSE of 0.0961, MAE of 0.2803, and Score of 297.3991, whereas the non-aligned model achieved an MSE of 0.1049, MAE of 0.2919, and Score of 321.70. These discrepancies likely arise from residual differences in label distributions and unmodeled domain-specific variations. While the alignment approach improves feature consistency across datasets, the remaining prediction error suggests that further optimization is needed to enhance model robustness.

\begin{table}[htbp]
\centering
\caption{Cross-domain Validation Experiment Results}
\vspace{3mm}
\label{tab:coral}
\footnotesize
\setlength{\tabcolsep}{4pt}
\renewcommand{\arraystretch}{0.5}
\begin{tabular}{@{}lSSS@{}}
\toprule
\rowcolor{gray!20}
\textbf{Model} & \textbf{MSE} & \textbf{MAE} & \textbf{Score} \\
\midrule
With CORAL & \textbf{0.0961} & \textbf{0.2803} & \textbf{297.3991} \\
Without CORAL & 0.1049 & 0.2919 & 321.7089 \\
\bottomrule
\end{tabular}
\vspace{3mm}
\begin{minipage}{\textwidth}
\footnotesize
\vspace{3mm}
\textbf{Note:} Bold values indicate the minimum MSE, MAE, and Score.
\end{minipage}
\end{table}

\subsection{Comparison with Baseline Methods}

To comprehensively evaluate the performance of CARLE, we conducted comparative experiments against several baseline methods, including CNN-LSTM~\cite{hu2024wear}, CNN-BiLSTM~\cite{guo2023cnn}, and MSIDIN~\cite{zhao2023multi}. For a fair comparison, all competing models were trained using feature vectors extracted by the proposed compact feature extractor framework. Additionally, hyperparameters for each method, including CARLE, were fine-tuned using Bayesian Optimization \cite{snoek2012practical} with 150 search trials to ensure optimal performance. Under training operating conditions, most state-of-the-art models were able to estimate RUL with reasonable accuracy based on MSE and MAE metrics. However, CARLE consistently outperformed all other methods across both datasets, with particularly significant improvements observed under unseen operating conditions. Detailed comparison metrics for the XJTU-SY and PRONOSTIA datasets are provided in Table~\ref{tab:compare_xjtu} and Table~\ref{tab:compare_pronostia}, respectively. To further interpret these results, we examined \textit{Bearing 3} under each operating condition from both datasets. 

For the XJTU-SY dataset:

\begin{itemize}
    \item \textbf{35Hz12kN:} CARLE demonstrated the best performance with an MSE of 0.00220, MAE of 0.04087 and Score of 199.61. In comparison, MSIDIN recorded an MSE of 0.04383, MAE of 0.17848 and Score of 494.269. CABiLSTM reached an MSE of 2.41237, MAE of 1.24057 and Score of 3667.2644, while CNN-LSTM exhibited the poorest accuracy with an MSE of 8.6898, MAE of 1.81207 and Score of 5917.488.

    \item \textbf{37.5Hz11kN:} CARLE maintained superior results with an MSE of 0.01407, MAE of 0.10697 and Score of 2798.9778. MSIDIN followed with an MSE of 0.08407, MAE of 0.24669 and Score 2528.81, CABiLSTM showed degraded performance with an MSE of 0.71245, MAE of 0.67876, Score 6417.187, and CNN-LSTM further deteriorated to an MSE of 1.19243, MAE of 0.63259 and Score of 6304.808.

    \item \textbf{40Hz10kN:} CARLE achieved an MSE of 0.03805, MAE of 0.15631 and Score of 530.86. MSIDIN yielded an MSE of 0.09673, MAE of 0.26873 and Score of 2094.007, CABiLSTM followed with an MSE of 1.07763, MAE of 0.88477, Score of 1746.2798, and CNN-LSTM recorded an MSE of 1.39763, MAE of 0.76697 and Score of 1562.1542.
\end{itemize}

For the PRONOSTIA dataset:

\begin{itemize}
    \item \textbf{100Hz4kN:} CARLE again delivered optimal results, achieving an MSE of 0.00029, MAE of 0.01312 and Score of 70.195. MSIDIN followed with an MSE of 0.00049, MAE of 0.01723 and Score of 1000.6890, while CABiLSTM recorded an MSE of 0.00268, MAE of 0.04036 and Score of 5860.68. Interestingly, CNN-LSTM attained an MSE of 0.00033 but slightly outperformed CARLE on MAE with a score of 0.01294 and Score of 1834.09.
    \item \textbf{100Hz4.2kN:} CARLE obtained an MSE of 0.00831, MAE of 0.07488 and Score of 80.114. MSIDIN slightly outperformed CARLE in all metrics, with MSE of 0.00606, MAE of 0.06360 and Score of 530.557. CABiLSTM trailed behind with an MSE of 0.01240, MAE of 0.09776 and Score of 3550.281, and CNN-LSTM significantly underperformed, with an MSE of 0.17432, MAE of 0.34082 and Score of 1312.410.
    \item \textbf{100Hz5kN :} CARLE achieved an MSE of 0.14152, MAE of 0.17514 and Score of 55.231. MSIDIN reported higher error values with an MSE of 0.15967, MAE of 0.35579 and Score of 64.344, while CABiLSTM showed substantial degradation, reaching an MSE of 2.18729, MAE of 1.2095 and Score of 233.21. CNN-LSTM also performed poorly, with an MSE of 1.07811, MAE of 0.84026 and Score of 150.869.
\end{itemize}

These findings reinforce CARLE's ability to generalize effectively across different operating environments and its superior accuracy in both seen and unseen conditions. Notably, even in scenarios where other methods perform competitively under trained settings, CARLE maintains a robust edge, particularly in generalization to unseen conditions, which is critical in real-world prognostics applications.

\begin{table}[htbp]
\centering
\caption{Comparison with SOTA (XJTU-SY)}
\vspace{3mm}
\label{tab:compare_xjtu}
\footnotesize
\setlength{\tabcolsep}{2pt}
\renewcommand{\arraystretch}{1}
\resizebox{\textwidth}{!}{%
\begin{tabular}{@{}llSSSSSSSSS@{}}
\toprule
\rowcolor{gray!20}
\multirow{2}{*}{\textbf{Bearing}} & \multirow{2}{*}{\textbf{Model}} &
\multicolumn{3}{c}{\textbf{35Hz12kN}} &
\multicolumn{3}{c}{\textbf{37.5Hz11kN}} &
\multicolumn{3}{c}{\textbf{40Hz10kN}} \\
\cmidrule(lr){3-5} \cmidrule(lr){6-8} \cmidrule(lr){9-11}
 & & \textbf{MSE} & \textbf{MAE} & \textbf{Score} & \textbf{MSE} & \textbf{MAE} & \textbf{Score} & \textbf{MSE} & \textbf{MAE} & \textbf{Score} \\
\midrule

\multirow{4}{*}{\textbf{Bearing 1}}
 & CARLE & \textbf{0.00345} & \textbf{0.05157} & \textbf{188.36298} & \textbf{0.03273} & \textbf{0.16070} & 2462.88 & \textbf{0.03314} & \textbf{0.15983} & 4009.317 \\
 & CNN-LSTM & 7.53585 & 1.68644 & 4212.5586 & 5.53985 & 1.19735 & 12144.208 & 15.12136 & 3.59279 & 54348.28 \\
 & CABiLSTM & 2.31171 & 1.23203 & 2790.2166 & 1.58179 & 0.90853 & 8101.4062 & 5.76011 & 2.28761 & 32350.693 \\
 & MSIDIN & 0.05074 & 0.17379 & 415.75772 & 0.08555 & 0.24885 & \textbf{2385.1533} & 0.08555 & 0.25344 & \textbf{3731.8108} \\
\midrule

\multirow{4}{*}{\textbf{Bearing 2}}
 & CARLE & \textbf{0.00183} & \textbf{0.03692} & \textbf{217.17297} & \textbf{0.00671} & \textbf{0.07308} & 991.1647 & \textbf{0.06673} & \textbf{0.22032} & 2266.9412 \\
 & CNN-LSTM & 2.76083 & 0.93285 & 2875.2263 & 4.41759 & 1.27952 & 4030.5173 & 0.83365 & 0.85786 & 6560.502 \\
 & CABiLSTM & 0.92567 & 0.70848 & 2034.5358 & 1.53518 & 0.96063 & 2848.8662 & 1.19016 & 1.03863 & 7994.697 \\
 & MSIDIN & 0.05763 & 0.17848 & 562.8828 & 0.12322 & 0.28668 & \textbf{916.3111} & 0.08829 & 0.25199 & \textbf{2094.0076} \\
\midrule

\multirow{4}{*}{\textbf{Bearing 3}}
 & CARLE & \textbf{0.00220} & \textbf{0.04087} & \textbf{199.6124} & \textbf{0.01407} & \textbf{0.10697} & 2798.9778 & \textbf{0.03085} & \textbf{0.15631} & \textbf{530.86194} \\
 & CNN-LSTM & 8.69865 & 1.81207 & 5917.488 & 1.19243 & 0.63258 & 6304.898 & 1.39763 & 0.76697 & 1562.1542 \\
 & CABiLSTM & 2.41237 & 1.24057 & 3667.2644 & 0.71245 & 0.67876 & 6417.187 & 1.07763 & 0.88477 & 1746.2798 \\
 & MSIDIN & 0.04383 & 0.15704 & 494.26984 & 0.08407 & 0.24669 & \textbf{2528.8157} & 0.09673 & 0.26387 & 554.0628 \\
\midrule

\multirow{4}{*}{\textbf{Bearing 4}}
 & CARLE & \textbf{0.01172} & \textbf{0.09653} & \textbf{292.51904} & \textbf{0.02149} & \textbf{0.12591} & 245.14308 & \textbf{0.03361} & \textbf{0.16221} & \textbf{2362.9336} \\
 & CNN-LSTM & 2.75389 & 0.81807 & 1996.5586 & 2.34120 & 0.92475 & 729.2322 & 0.85732 & 0.66800 & 5408.3696 \\
 & CABiLSTM & 1.21288 & 0.81330 & 1857.263 & 1.11647 & 0.87074 & 651.8345 & 0.92342 & 0.82073 & 6554.092 \\
 & MSIDIN & 0.08448 & 0.24985 & 592.63196 & 0.08910 & 0.25441 & \textbf{198.84254} & 0.11538 & 0.27989 & 2404.415 \\
\midrule

\multirow{4}{*}{\textbf{Bearing 5}}
 & CARLE & \textbf{0.00465} & \textbf{0.05938} & \textbf{92.718994} & \textbf{0.01373} & \textbf{0.09903} & 1890.4968 & \textbf{0.09625} & \textbf{0.26167} & \textbf{190.29407} \\
 & CNN-LSTM & 3.28659 & 1.36531 & 1333.5254 & 0.78787 & 0.78197 & 4740.767 & 1.34206 & 1.11198 & 669.8485 \\
 & CABiLSTM & 2.04337 & 1.19535 & 1150.7625 & 0.52122 & 0.61763 & 3721.561 & 1.00430 & 0.95609 & 572.3747 \\
 & MSIDIN & 0.06368 & 0.19819 & 204.28513 & 0.11293 & 0.28494 & \textbf{1821.144} & 0.35019 & 0.53957 & 319.7993 \\
\bottomrule
\end{tabular}}
\begin{minipage}{\textwidth}
\footnotesize
\vspace{3mm}
\textbf{Note:} Bold values indicate the minimum MSE, MAE, and Score for each bearing-condition combination across all models.
\end{minipage}
\end{table}

\begin{table}[htbp]
\centering
\caption{Comparison with SOTA (PRONOSTIA)}
\vspace{3mm}
\label{tab:compare_pronostia}
\footnotesize
\setlength{\tabcolsep}{2pt}
\renewcommand{\arraystretch}{1}
\resizebox{\textwidth}{!}{%
\begin{tabular}{@{}llSSSSSSSSSS@{}}
\toprule
\rowcolor{gray!20}
\multirow{2}{*}{\textbf{Bearing}} & \multirow{2}{*}{\textbf{Model}} &
\multicolumn{3}{c}{\textbf{100Hz4kN}} &
\multicolumn{3}{c}{\textbf{100Hz4.2kN}} &
\multicolumn{3}{c}{\textbf{100Hz5kN}} \\
\cmidrule(lr){3-5} \cmidrule(lr){6-8} \cmidrule(lr){9-11}
 & & \textbf{MSE} & \textbf{MAE} & \textbf{Score} & \textbf{MSE} & \textbf{MAE} & \textbf{Score} & \textbf{MSE} & \textbf{MAE} & \textbf{Score} \\
\midrule

\multirow{4}{*}{\textbf{Bearing 1}}
 & CARLE & \textbf{0.00017} & \textbf{0.00890} & \textbf{74.451} & \textbf{0.00687} & \textbf{0.06874} & \textbf{37.435} & \textbf{0.03073} & \textbf{0.15232} & \textbf{24.860} \\
 & CNN-LSTM & 0.00060 & 0.01944 & 4444.265 & 0.20664 & 0.39369 & 249.752 & 0.01941 & 0.12100 & 197.559 \\
 & CABiLSTM & 0.00130 & 0.02515 & 6380.369 & 0.01075 & 0.08844 & 477.703 & 0.02536 & 0.13877 & 281.832 \\
 & MSIDIN & 0.00055 & 0.01640 & 1488.338 & 0.00720 & 0.07223 & 156.422 & 0.03786 & 0.17235 & 88.091 \\
\midrule

\multirow{4}{*}{\textbf{Bearing 2}}
 & CARLE & \textbf{0.00289} & \textbf{0.04268} & \textbf{32.028} & \textbf{0.00406} & \textbf{0.05340} & \textbf{25.583} & \textbf{0.04126} & \textbf{0.17514} & \textbf{54.820} \\
 & CNN-LSTM & 0.01822 & 0.10301 & 322.214 & 0.09584 & 0.26255 & 340.356 & 0.06594 & 0.22073 & 412.641 \\
 & CABiLSTM & 0.01010 & 0.08341 & 602.274 & 0.00706 & 0.06948 & 1421.654 & 0.02629 & 0.14069 & 1623.838 \\
 & MSIDIN & 0.00643 & 0.06490 & 159.866 & 0.00403 & 0.05156 & 154.332 & 0.04196 & 0.17951 & 330.132 \\
\midrule

\multirow{4}{*}{\textbf{Bearing 3}}
 & CARLE & \textbf{0.00029} & \textbf{0.01312} & \textbf{70.194} & \textbf{0.00831} & \textbf{0.07488} & \textbf{80.114} & \textbf{0.141255} & \textbf{0.17514} & \textbf{55.231} \\
 & CNN-LSTM & 0.00033 & 0.01294 & 1834.090 & 0.17432 & 0.34082 & 1312.410 & 1.078112 & 0.845026 & 150.869 \\
 & CABiLSTM & 0.00094 & 0.02538 & 5860.680 & 0.01240 & 0.09776 & 3550.281 & 2.1872909 & 1.20955 & 233.210 \\
 & MSIDIN & 0.00049 & 0.01723 & 1000.680 & 0.00606 & 0.06360 & 530.557 & 0.159579 & 0.355762 & 64.344 \\
\midrule

\multirow{4}{*}{\textbf{Bearing 4}}
 & CARLE & \textbf{0.00052} & \textbf{0.01635} & \textbf{39.394} & \textbf{0.01035} & \textbf{0.07616} & \textbf{41.840} & \text{--} & \text{--} & \text{--} \\
 & CNN-LSTM & 0.00237 & 0.03209 & 496.519 & 0.02068 & 0.12315 & 343.611 & \text{--} & \text{--} & \text{--} \\
 & CABiLSTM & 0.00268 & 0.04036 & 809.504 & 0.01279 & 0.09542 & 490.663 & \text{--} & \text{--} & \text{--} \\
 & MSIDIN & 0.00134 & 0.02676 & 228.168 & 0.01029 & 0.08145 & 132.623 & \text{--} & \text{--} & \text{--} \\
\midrule

\multirow{4}{*}{\textbf{Bearing 5}}
 & CARLE & \textbf{0.00264} & \textbf{0.03956} & \textbf{79.403} & \textbf{0.00711} & \textbf{0.06936} & \textbf{102.632} & \text{--} & \text{--} & \text{--} \\
 & CNN-LSTM & 0.00885 & 0.07636 & 1135.705 & 0.08058 & 0.23612 & 951.493 & \text{--} & \text{--} & \text{--} \\
 & CABiLSTM & 0.00999 & 0.08285 & 2605.018 & 0.01398 & 0.10261 & 3279.224 & \text{--} & \text{--} & \text{--} \\
 & MSIDIN & 0.00605 & 0.06274 & 442.709 & 0.00878 & 0.07688 & 448.584 & \text{--} & \text{--} & \text{--} \\
\midrule

\multirow{4}{*}{\textbf{Bearing 6}}
 & CARLE & \textbf{0.02014} & \textbf{0.12283} & \textbf{110.142} & \textbf{0.07713} & \textbf{0.23869} & \textbf{27.673} & \text{--} & \text{--} & \text{--} \\
 & CNN-LSTM & 0.03497 & 0.13915 & 1099.476 & 0.20729 & 0.39659 & 332.571 & \text{--} & \text{--} & \text{--} \\
 & CABiLSTM & 0.03064 & 0.15598 & 2633.624 & 0.03074 & 0.15201 & 447.319 & \text{--} & \text{--} & \text{--} \\
 & MSIDIN & 0.05942 & 0.20851 & 619.222 & 0.03314 & 0.15922 & 106.045 & \text{--} & \text{--} & \text{--} \\
\midrule

\multirow{4}{*}{\textbf{Bearing 7}}
 & CARLE & \textbf{0.00799} & \textbf{0.06877} & \textbf{112.334} & 0.00495 & \textbf{0.06176} & \textbf{13.302} & \text{--} & \text{--} & \text{--} \\
 & CNN-LSTM & 0.03480 & 0.13173 & 878.371 & 0.06695 & 0.18207 & 63.732 & \text{--} & \text{--} & \text{--} \\
 & CABiLSTM & 0.01221 & 0.08969 & 1799.508 & 0.00929 & 0.08194 & 166.860 & \text{--} & \text{--} & \text{--} \\
 & MSIDIN & 0.00905 & 0.07806 & 362.584 & \textbf{0.00355} & 0.05256 & 33.425 & \text{--} & \text{--} & \text{--} \\
\bottomrule
\end{tabular}}
\begin{minipage}{\textwidth}
\footnotesize
\vspace{3mm}
\textbf{Note:} Bold values indicate the minimum MSE, MAE, and Score for each bearing–condition combination.
\end{minipage}
\end{table}

\subsection{Explanations}
Higher accuracy in an AI system does not necessarily mean its predictions reflect real-world outcomes \cite{bontempi2023between}. This makes it essential to direct explainable AI (XAI) efforts toward PHM systems, particularly for remaining useful life (RUL) analysis of mechanical components, where unexpected failures can cause major operational disruptions. In this study, we applied Local Interpretable Model-Agnostic Explanations (LIME) \cite{ribeiro2016should} and Shapley Additive Explanations (SHAP) \cite{lundberg2017unified} to interpret model predictions.  

\begin{figure}[ht!]
\centering
\includegraphics[width=0.95\textwidth]{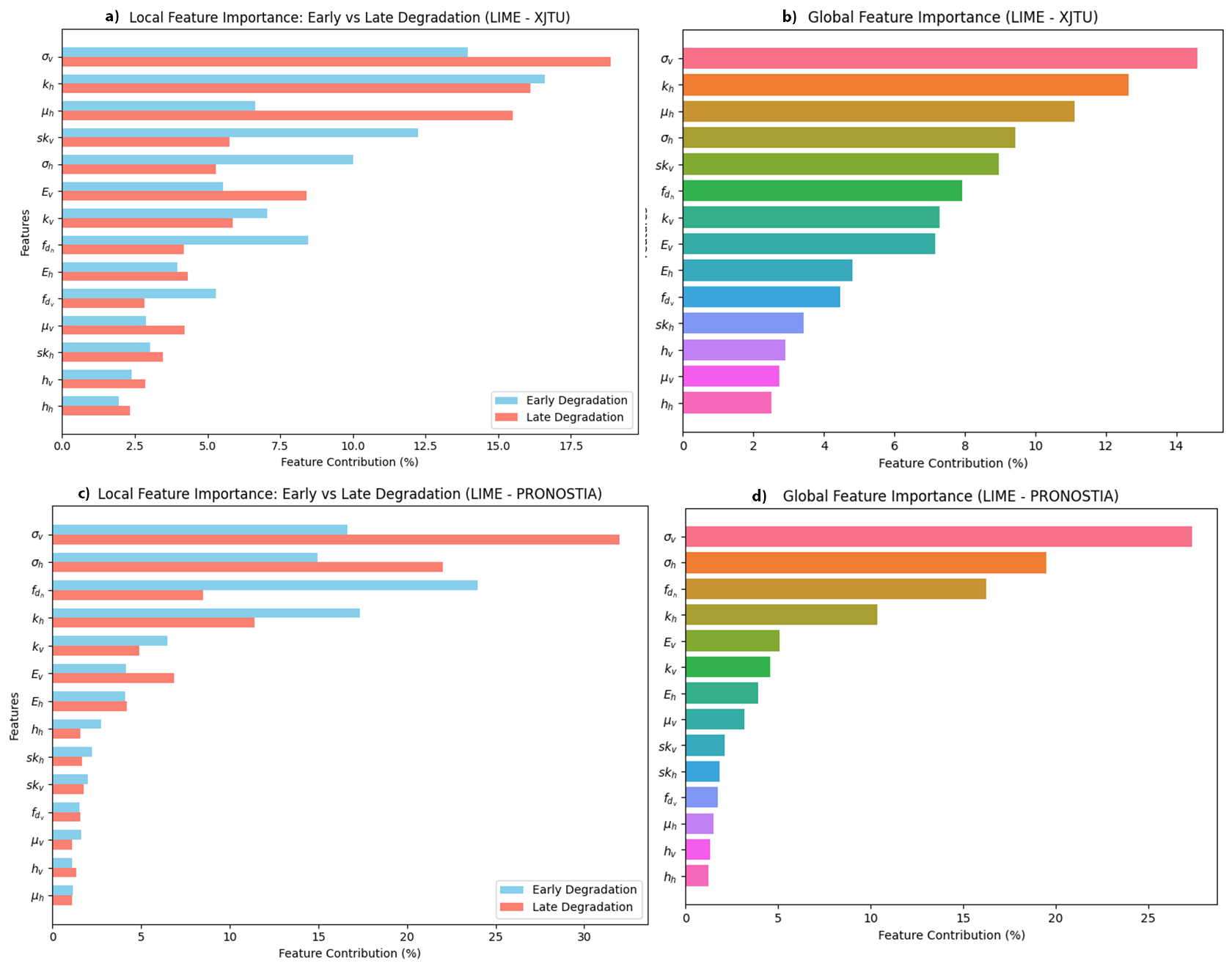}
\caption{LIME explanation for \textbf{a)} XJTU-SY; \textbf{b)} for PRONOSTIA.}
\label{fig:lime}
\end{figure}

We selected two test points, one from the early degradation stage and one from the late degradation stage, to examine which features contribute most during fault development. Figure~\ref{fig:lime}(a,c) shows local explanations for XJTU and PRONOSTIA. In the early stage, $\sigma_v$ played the most significant role in predictions, followed by $k_v$. This suggests that early degradation is primarily reflected in increased vibration variability and subtle distributional changes such as heavier tails. In practice, these effects correspond to small surface defects or early spalls on the bearing raceway that disturb the signal but do not yet dominate its frequency content.  

As degradation progressed, the influence of $\sigma$ and $\mu$ increased substantially, with \(k\) becoming the second most important feature. These variables capture more pronounced shifts in the vibration component and distributional asymmetry, which in real-world terms correspond to advanced fault development. At this stage, cracks expand, spalls deepen, and defect impacts become stronger and more asymmetric, producing larger and more irregular vibrations that are easier to isolate. To generate global insights, local explanations were aggregated to identify the vibration characteristics most critical to bearing degradation and RUL estimation. Results (Figure~\ref{fig:lime})(b,d) show that both the XJTU-SY and PRONOSTIA models rely heavily on $\sigma$, a measure of signal variability. This finding aligns with the physics of bearing failure, where increased variability often signals instability caused by defects such as looseness, contamination, or misalignment. The models also prioritize $f_d$ components, which capture dominant frequency shifts associated with localized faults such as inner and outer race cracks, spalling, or lubrication deficiencies. In contrast, $h$ contributes minimally, likely because fragmenting signals into shorter time windows reduces sensitivity to this global feature.  

\begin{figure}[ht!]
\centering
\includegraphics[width=0.95\textwidth]{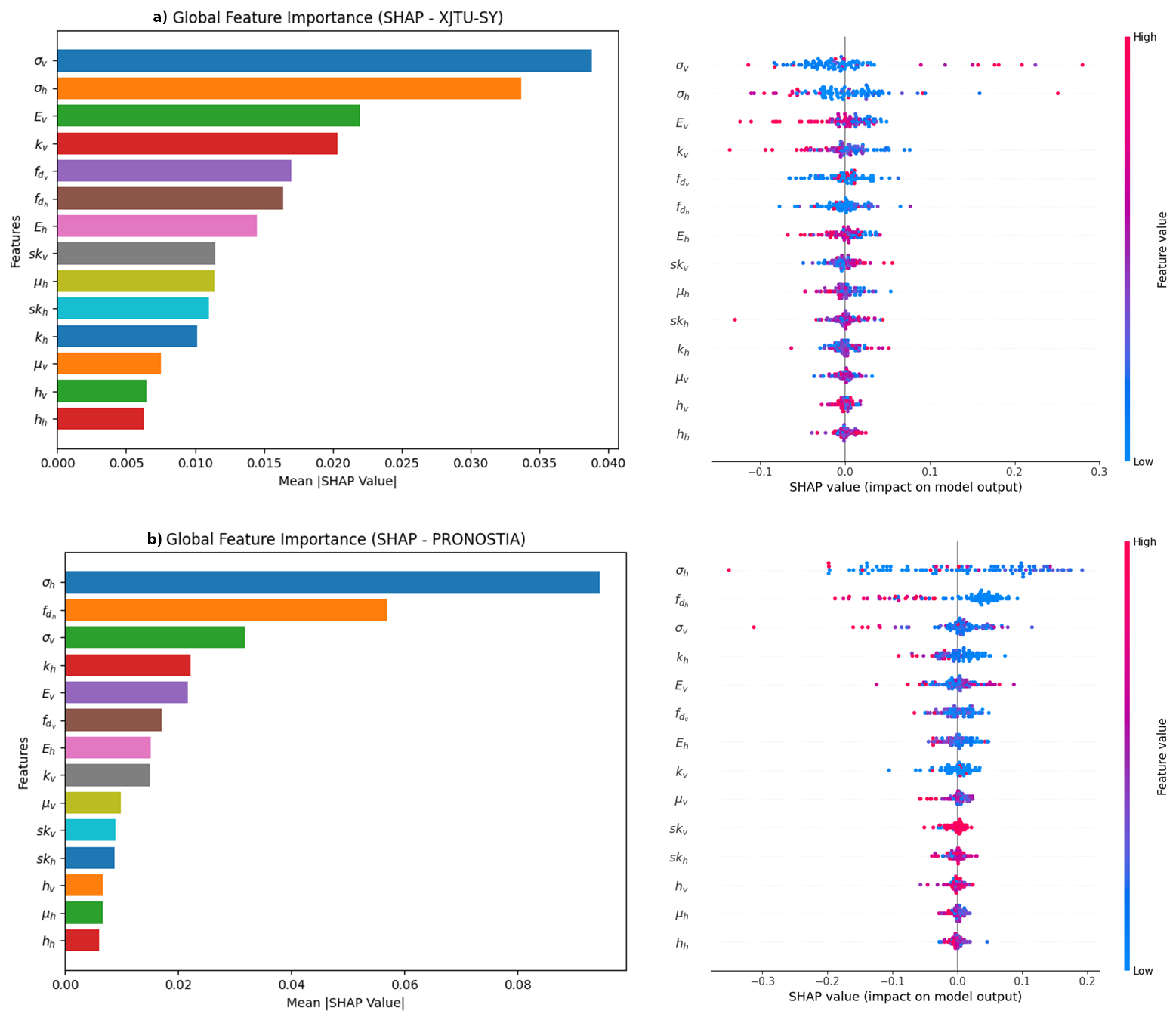}
\caption{SHAP explanation for \textbf{a)} XJTU-SY; \textbf{b)} for PRONOSTIA.}
\label{fig:shap}
\end{figure}

SHAP analysis (Figure~\ref{fig:shap}) confirms these findings and adds nuance. $\sigma$ has the largest absolute impact, indicating that overall \(\sigma\) is the most reliable predictor of degradation. \(f_d\) components follow closely, reflecting the model's ability to capture fault-specific signatures. $E$ features also contribute significantly, linking directly to failure mechanisms such as spalling progression, crack propagation, and lubrication breakdown. By contrast, $h$ remains the least influential feature, confirming that short window fragmentation reduces its predictive power. 

This detailed feature-level interpretation shows that CARLE not only produces accurate RUL predictions but does so in a way that reflects the underlying physical processes of bearing degradation, increasing both trust and applicability in high-risk industrial settings.

\section{Conclusion}
This research proposes a comprehensive RUL estimation system for rolling-element bearings. The system comprises three key components: a compact time‒frequency feature extraction framework, an AI framework (CARLE), and XAI explanations. The feature extractor framework includes a complete algorithm to transform non-stationary vibrational signals into a set of time‒frequency features using CWT. It also incorporates a Gaussian noise filter to eliminate signal perturbations and short-term fluctuations. The CARLE AI framework comprises four blocks: Res-CNN captures spatial degradation trends from the input feature set; Res-RNN captures temporal degradation trends, learning long-term time dependencies; Linear block identifies patterns within these dependencies to produce a logit vector; finally, RFR predicts the final RUL. This ensemble approach, combining deep learning and traditional machine learning methods, enhances robustness and generalization, allowing the system to adapt effectively from one working condition to unseen conditions. We evaluated the trustworthiness of the AI framework using aggregated LIME and SHAP. The analysis revealed that CARLE heavily relies on $\sigma$ features, which indicate that unstable faults such as looseness or contamination cause erratic behavior. The analysis also revealed that both models heavily rely on $f_d$, which is an indicator of localized defects, including inner and outer race cracks, looseness, and lubrication failures. Additionally, SHAP suggests that $E$ features are also important, as they indicate mechanical stress, friction, and surface defects. Other factors contribute but are less significant, confirming the system’s reliability. We validated the proposed framework using the XJTU-SY and PRONOSTIA benchmark datasets.

\subsection{Future Work}
While the findings of this research are promising, there is still room for improvement. We observed that CARLE struggles with early fault detection (see Figure \ref{fig:ablation_exp}(a(xii-xiii), b(xvii))). Early degradation detection could be improved by incorporating a physics-guided loss to better capture subtle changes in the initial stages of degradation. Cross-domain validation experiments indicate that further hyperparameter tuning could enhance CARLE's generalization performance. Another possible mitigation is to incorporate domain-adaptive training or fine-tuning on the target dataset to better capture domain-specific label distributions. Furthermore, in real-world scenarios, run-to-failure datasets are often unavailable. Implementing CARLE in a transfer learning configuration with incomplete run-to-failure data is also a promising direction for future research.

\section*{Conflict of Interest}
The authors declare that they have no conflicts of interest to disclose.

\section*{Ethics Approval}
This study was conducted in accordance with ethical standards.

\section*{Funding}
The research did not receive any funding from any organization.

\section*{Data Availability}
The code is available at \texttt{https://github.com/itxwaleedrazzaq/PhDCode.git}.

\section*{Authors Contribution}
\textbf{Waleed Razzaq:} Conceptualization, Methodology, Data Curation, Writing- Original draft preparation. \textbf{Yun-Bo Zhao}: Supervision, Writing- Reviewing.

\section*{Acknowledgment}
This research was supported by the CAS-ANSO Scholarship. We acknowledge the intellectual and material contributions of the University of Science and Technology of China (USTC) and the Alliance of International Science Organizations (ANSO).

\section*{Human/Animal Participation}
No human or animal participation is involved in this research.

\bibliographystyle{unsrt}
\bibliography{references}

\newpage

\appendix
\section*{Appendix}
\section{Preliminaries}
In this section, we provide an overview of some building blocks of our proposed framework.
\subsection{Gaussian Filter}
The Gaussian filter \(G(x) \) is a smoothing filter commonly used to reduce noise, smooth data, and extract trends from non-stationary signals, which are crucial in predicting the RUL. It applies a weighted averaging operation to the signal, ensuring that values closer to the center of the filter contribute more to the result than those farther away. The mathematical expression of the Gaussian function is given by:
\begin{align}
G(x) = \frac{1}{\sqrt{2\pi\sigma_g^2}} e^{-\frac{x_d^2}{2\sigma^2}}
\end{align}
where \(x_d \) is the distance from the center of the filter. \(\sigma_g \) is the standard deviation of the Gaussian distribution, which controls the width of the Gaussian curve and determines the degree of smoothness.

\subsection{Continuous Wavelet Transform}

The Continuous Wavelet Transform (CWT) is a powerful mathematical tool that decomposes a time-varying signal into highly localized oscillations called wavelets, providing better time‒frequency analysis. The CWT uses basis functions that are scaled and shifted versions of the time-localized wavelet, enabling the creation of a time-frequency representation of a signal with excellent localization in both time and frequency. The mathematical expression of the CWT is as follows:
\begin{align}
\Gamma(a,b) = \int_{-\infty}^{\infty} I(t) \psi^* \left(\frac{t - b}{a} \right) dt
\end{align}

where \(\Gamma(a,b) \) represents the wavelet coefficients at scale \(a \) and translation \(b \), \(I(t) \) represents the nonstationary signal, and \(\psi(t) \) represents the mother wavelet function. We selected the Morlet wavelet \cite{farge1992wavelet} as the mother wavelet for time-frequency representation (TFR) extraction due to its similarity to the bearing impulse response \cite{zhu2018estimation} and its favorable trade-off between time and frequency resolution. In particular, its frequency resolution improves at higher values of \(a\), while the time resolution improves at lower values \cite{lin2000feature}. The Morlet wavelet is defined as a sinusoidal function modulated by a Gaussian envelope with a central frequency \(f_c \) and is given by:
\begin{align}
\psi(t) = e^{\frac{if_c t}{2\pi}} e^{-t^2/2}
\end{align}

\subsection{Long Short-Term Memory (LSTM)}

\begin{figure}[h!]
\centering
\includegraphics[width=0.6\textwidth]{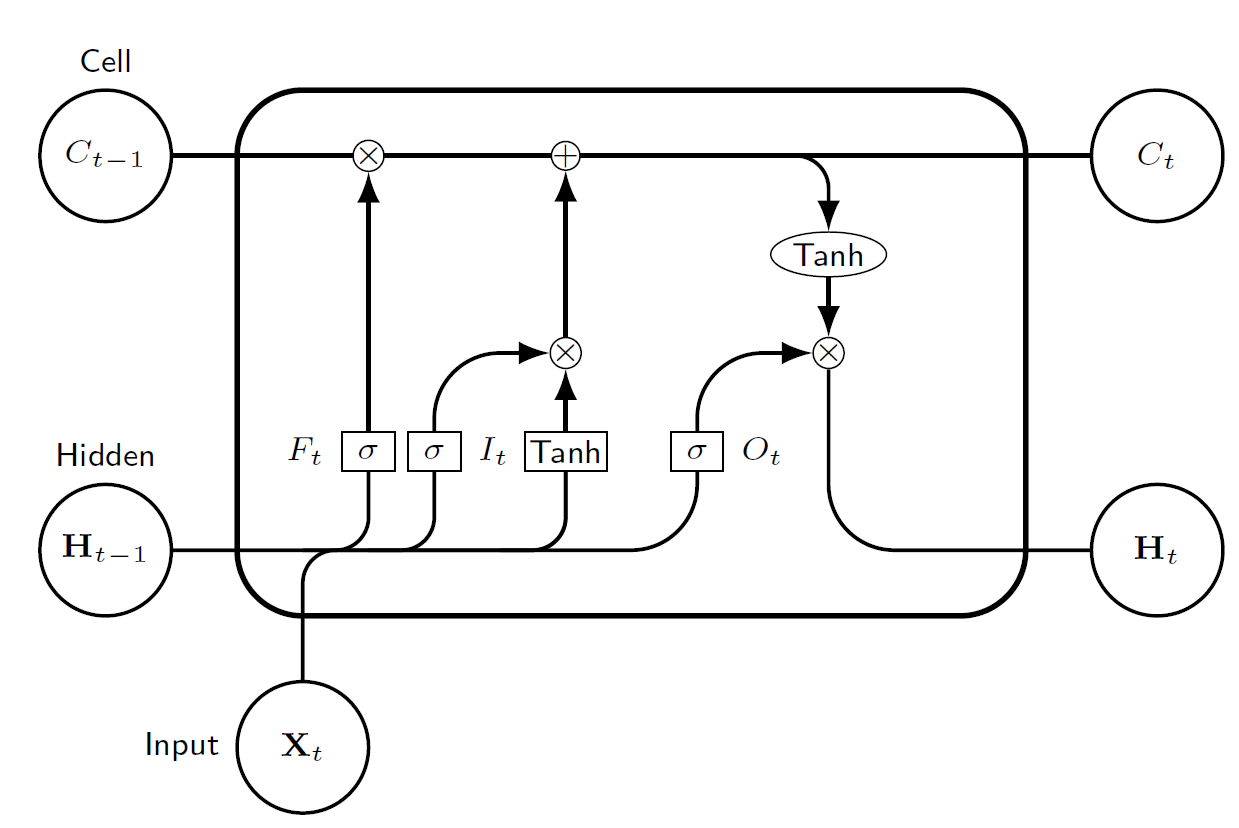}
\caption{Structure of the LSTM network.}
\label{fig1}
\end{figure}

The LSTM network is a class of deep recurrent networks designed to capture long-term time dependencies from data. LSTM utilizes specialized gates, i.e., an input gate $I_t$, a forget gate $F_t$, and an output gate $O_t$, to regulate the flow of information, allowing selective retention and forgetting of information. This ability makes LSTM ideal for modeling time series data that exhibit long-term dependencies such as the gradual degradation of rolling element bearings, providing a more accurate RUL estimation \cite{hochreiter1997long}. The structure of an LSTM network is shown in Figure \ref{fig1}, and the output of an LSTM network can be mathematically modeled as:
\begin{align}
\mathbf{H}_t = \mathcal{NN}(\mathbf{I}_t, \mathbf{H}_{t-1}) =
\begin{cases}
    C_t = \phi(\mathbf{W}_g [\mathbf{H}_{t-1}, \mathbf{X}_t] + \mathbf{b}_g) \\
    I_t = \sigma(\mathbf{W}_i [\mathbf{H}_{t-1}, \mathbf{X}_t] + \mathbf{b}_i) \\
    F_t = \sigma(\mathbf{W}_f [\mathbf{H}_{t-1}, \mathbf{X}_t] + \mathbf{b}_f) \\
    O_t = \sigma(\mathbf{W}_o [\mathbf{H}_{t-1}, \mathbf{X}_t] + \mathbf{b}_o) \\
    \mathbf{S}_t = C_t \odot X_t + \mathbf{S}_{t-1} \odot F_t \\
    \mathbf{H}_t = O_t \odot \phi(\mathbf{S}_t)
\end{cases}
\end{align}

\subsection{Random Forest Regressor}

Random Forest Regression (RFR) is a supervised learning algorithm that employs an ensemble learning method for regression tasks based on the bagging technique. In RFR, the trees operate in parallel, meaning that there is no interaction between them during the training process. Each tree is trained on a random subset of the features, and the final prediction is obtained by averaging the outputs of all the trees \cite{segal2004machine}. We chose RFR for its accuracy, robustness, and ability to handle nonlinear relationships effectively in data, making it particularly suitable for RUL estimation, where complex interactions and temporal patterns are crucial. A schematic diagram of RFR is shown in Figure \label{fig2}.

\begin{figure}[h!]
\centering
\includegraphics[width=0.6\textwidth]{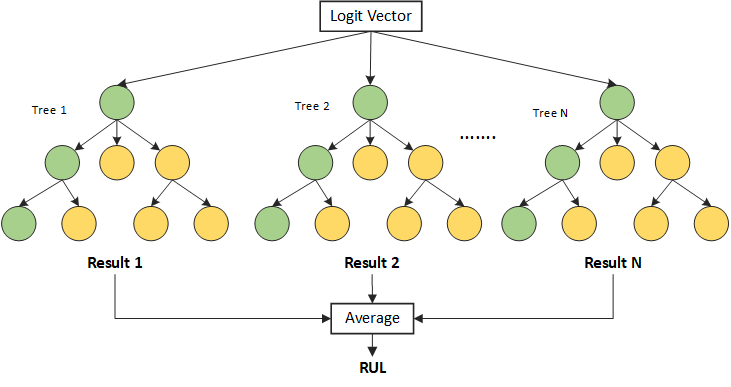}
\caption{Structure of the RFR algorithm.}
\label{fig2}
\end{figure}

\section{Implementation}
In this section, we provide the hyperparameters for both XJTU-SY and PRONOSTIA and the training regularization and optimizations that we use in our implementation.

\subsection{Training Setup}
We trained our CARLE on an Intel Core i5-7200U with 16 GB RAM and no GPU. The model was implemented in Python 3.10 using Tensorflow 2.18. Due to computational limitations and to make training efficient, we applied various optimizations to improve training efficiency. To ensure that the model converges to the best possible solution despite hardware constraints, we incorporate several callbacks: \textit{ResetStateCallback} to reset model states between epochs, \textit{EarlyStopping} to halt training if validation loss stagnates for multiple epochs, \textit{ReduceLROnPlateau} to adjust the learning rate on MSE dynamically, and \textit{ModelCheckpoint} to save the best training weights. These optimizations collectively enhance both training efficiency and model performance.

\begin{algorithm}[ht!]
\caption{Training and Testing of CARLE}\label{algo:training}
\begin{algorithmic}[1]
\Require Features vector ($I_{fv}$), RUL labels ($Y$), CARLE: $f(x,w) \to y$, Loss: $L(y,\hat{y}) \to R$, batch size $k$, Number of trees ($n_{trees}$)
\State Initialize weights $w$
\State $l_{min} \gets \infty$
\State Initialize empty forest: $Trees \gets \{\}$
\Procedure{Deep Neural Network}{}
\For{$e = 1 \ldots max_{EPOCH}$}
    \For{$i = 1 \ldots [\frac{X}{k}]$}
        \State $(x, y)$ is the batch size of $k$ from $(I_{fv}, Y)$
        \State $w \gets w_{t-1} - \frac{\eta}{E[g^2]_t}\frac{\partial C}{\partial w}$ \Comment{Root mean square prop}
    \EndFor
    \State Compute loss metrics:
        \State $l_e = \sqrt{\sum_{i=1}^n (y - \hat{y})^2}$ \Comment{Root mean square error}
        \State $mae = \frac{1}{n} \sum_{i=1}^{n} \left| y_i - \hat{y}_i \right|$ \Comment{Mean absolute error}
    \If{$l_e < l_{min}$}
        \State $l_{min} \gets l_e$
        \State $w_{best} \gets w_e$
    \EndIf
\EndFor
\State Compute Logit vector: $I_{lv} \gets \textit{CARLE}(I_{fv})$ \Comment{Output from CARLE}
\EndProcedure
\Procedure{Random Forest Regression}{}
\For{$e = 1 \ldots n_{trees}$}

    \State Initialize decision tree $T_e$ with $max_{feat}$
    \State Train $T_e$ on $((I_{lv}, Y))$: $T_e \gets \textit{fit}((I_{lv}, Y))$
    \State Add trained tree to forest: $Trees \gets Trees \cup \{T_e\}$
\EndFor

\State Compute training predictions: $\hat{Y} \gets \frac{1}{n_{trees}} \sum_{e=1}^{n_{trees}} T_e(I_{lv})$
\State Compute loss metrics:
    \State $MSE = \sqrt{\frac{1}{n} \sum_{i=1}^{n} (y_i - \hat{y}_i)^2}$ \Comment{Root mean square error}
    \State $MAE = \frac{1}{n} \sum_{i=1}^{n} \left| y_i - \hat{y}_i \right|$ \Comment{Mean absolute error}
\If{$MSE < l_{min}$}
    \State $l_{min} \gets MSE$
    \State $Best_{Forest} \gets Trees$
\EndIf

\State $rul \gets \hat{Y}$ \Comment{Output from Random Forest}
\EndProcedure

\State \textbf{return} $w_{best}, Best_{Forest}, rul$

\end{algorithmic}
\end{algorithm}

\begin{table}[htbp]
    \centering
    \caption{Hyperparameter comparison of CARLE (XJTU-SY vs PRONOSTIA)}
    \vspace{0.3 cm}
    \begin{tabular}{@{}l|lll@{}}
    \toprule
    \textbf{Block} & \textbf{Hyperparameter} & \textbf{XJTU-SY} & \textbf{PRONOSTIA} \\ \midrule

    \multirow{8}{*}{\textbf{Res-CNN}}
    & CNN Layers & 4 & 4 \\
    & CNN Filters & [256, 256, 128, 64] & [64, 64, 32, 32] \\
    & Kernel Sizes & [3, 3, 2, 2] & [3, 3, 2, 2] \\
    & Padding & Same & Same \\
    & Regularization ($\lambda$) & 0.005 & 0.005 \\
    & Activation & ReLU & ReLU \\
    & Pooling Size & 1 (MaxPooling1D) & 1 (MaxPooling1D) \\
    & Residual Connections & Applied & Applied \\
    & Multi-Head Attention & 8 Heads, 64 Dim & 8 Heads, 64 Dim \\ \midrule

    \multirow{6}{*}{\textbf{Res-LSTM}}
    & LSTM Layers & 2 & 2 \\
    & LSTM Units & [64,64] & [64,64] \\
    & Statefulness & False & False \\
    & Return Sequences & True & True \\
    & Residual Connections & Applied & Applied \\
    & Multi-Head Attention & 8 Heads, 64 Dim & 8 Heads, 64
    Dim \\
    & Flatten Layer & Applied & Applied \\
    \midrule

    \multirow{3}{*}{\textbf{Linear}}
    & Linear Layers & 3 & 3 \\
    & Linear Units & [128, 64, 32] & [64, 48, 32] \\
    \midrule

    \multirow{1}{*}{\textbf{Random Forest Regressor (RFR)}}
    & No. of trees &800  &800 \\
    \bottomrule
    \end{tabular}
    \label{tab:combined}
\end{table}
\begin{figure}[htbp]
\centering
\includegraphics[width=0.75\textwidth]{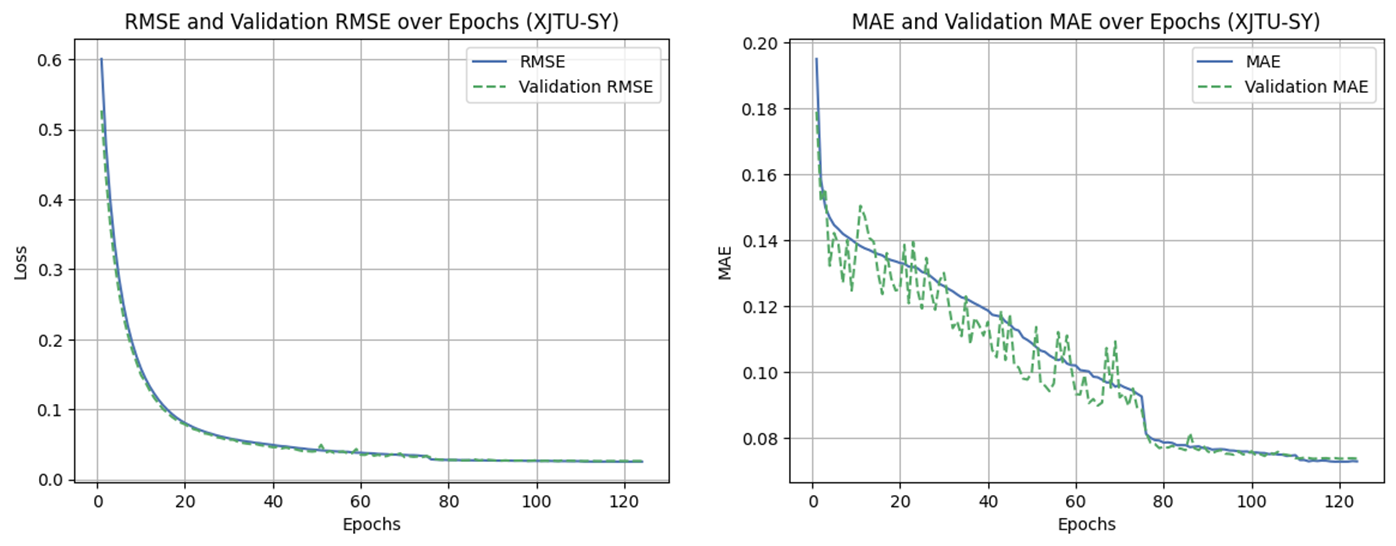}
\caption{Training statistics of CARLE (XJTU-SY).}
\label{fig:stat_xjtu}
\end{figure}

\begin{figure}[htbp]
\centering
\includegraphics[width=0.75\textwidth]{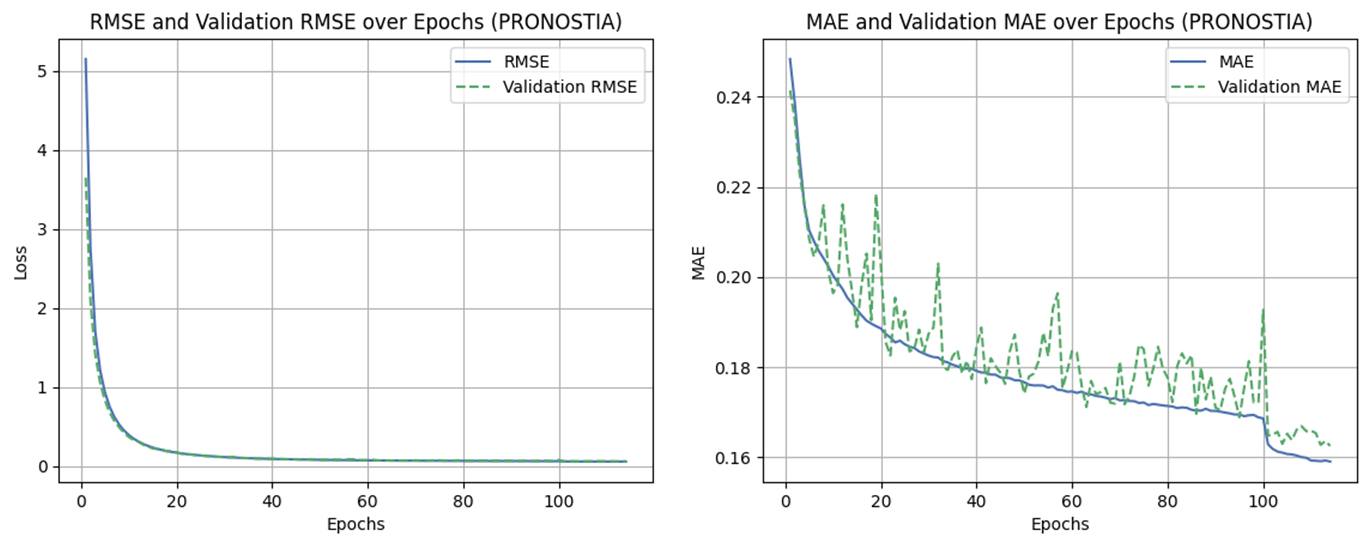}
\caption{Training statistics of CARLE (PRONOSTIA).}
\label{fig:stat_pronostia}
\end{figure}

\begin{figure}[htbp]
\centering
\includegraphics[width=0.65\textwidth]{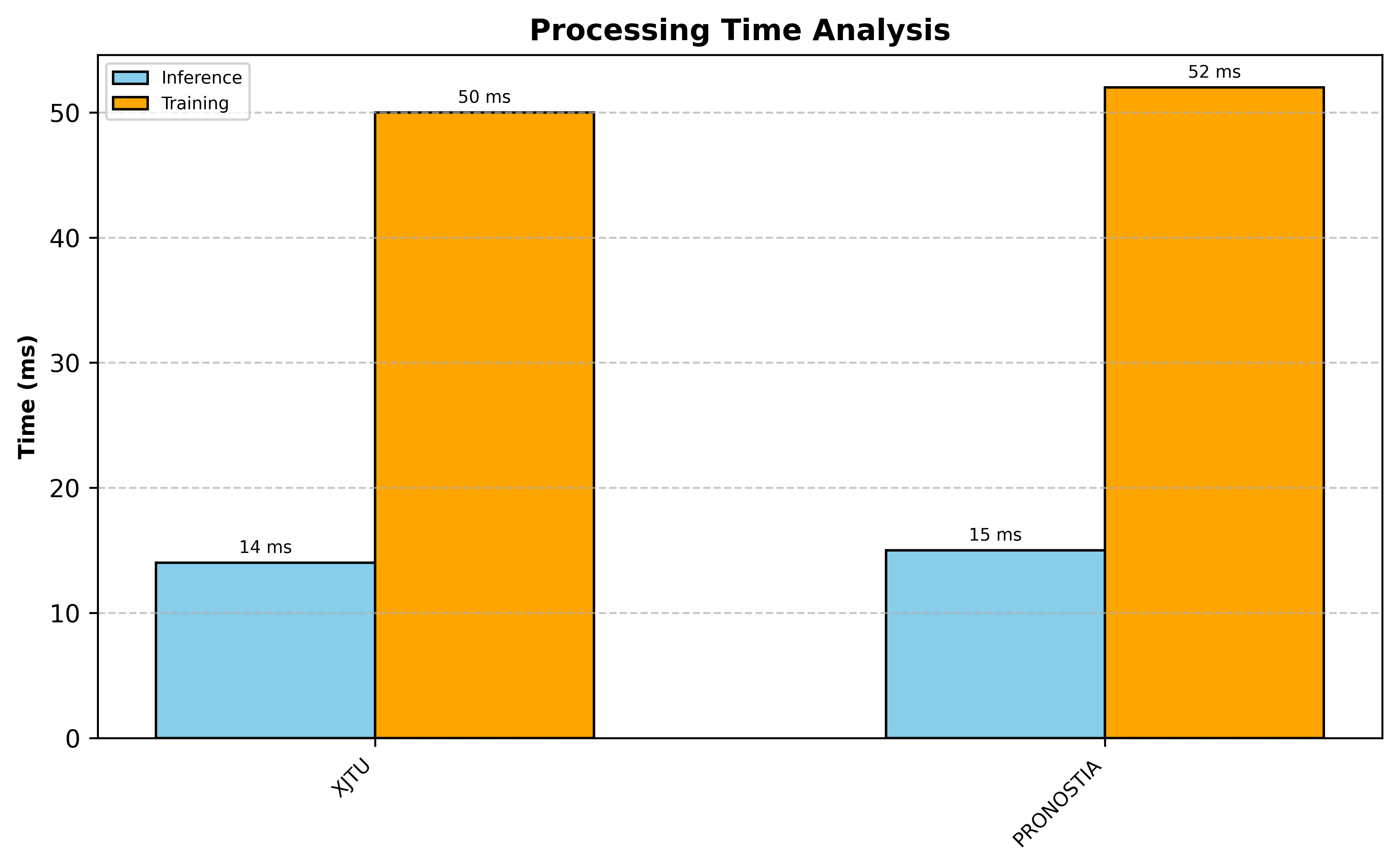}
\caption{Processing Time Analysis (XJTU-SY vs. PRONOSTIA).}
\label{fig:process}
\end{figure}

\subsection{Training and Testing of CARLE}
The training and testing procedure for CARLE involves two phases: training the deep neural network and training the random forest regression model. During the first phase, the model is optimized via batch updates and the MSE loss function to learn the relationships between the input features and RUL labels in a supervised manner. The output, a logit vector, is then used to train an RFR consisting of multiple decision trees. The MSE and MAE performance metrics are used throughout the training process to evaluate and select the best model. The trained neural network and RFR are applied to unseen data to predict the RUL during testing. The complete algorithm is detailed in Algorithm \ref{algo:training}, and the model parameters for XJTU-SY and PRONOSTIA are detailed in Table \ref{tab:combined}. The training statistics for XJTU-SY and PRONOSTIA are provided in Figure \ref{fig:stat_xjtu} and Figure \ref{fig:stat_pronostia}, respectively Time processing time analysis for both XJYU-SY and PRONOSTIA datasets are provided in Figure \ref{fig:process}. Both achieved nearly identical training and inference time in a moderate training setup. On low-end hardware, these processing times suggest that while training may be slower, the inference step, critical for real-time localized prognostics, remains feasible, as the model’s small size and low computational complexity enable fast forward passes even without high-end resources.

\end{document}